\documentclass{article} 
\usepackage{arxiv,times}

\usepackage{amsmath,amsfonts,bm}

\def\eqref#1{equation~\ref{#1}}

\def\1{\bm{1}}

\DeclareMathAlphabet{\mathsfit}{\encodingdefault}{\sfdefault}{m}{sl}
\SetMathAlphabet{\mathsfit}{bold}{\encodingdefault}{\sfdefault}{bx}{n}

\usepackage{hyperref}
\usepackage{url}
\usepackage[none]{hyphenat}

\usepackage{enumitem}
\usepackage{tikz}
\usetikzlibrary{arrows.meta,backgrounds,positioning}
\usepackage{xcolor}
\usepackage{float}
\usepackage{booktabs}
\usepackage{caption}
\usepackage[most]{tcolorbox}

\newtcolorbox{theorybox}[1][]{
    enhanced,
    breakable,
    colback=gray!5,
    colframe=gray!45,
    boxrule=0.4pt,
    arc=2pt,
    left=6pt,
    right=6pt,
    top=5pt,
    bottom=5pt,
    fonttitle=\bfseries,
    #1
}

\usepackage{microtype}
\title{Vmem-$\varphi$: Low-Compute Out-of-Distribution Detection in Spiking Neural Networks from Membrane-Potential Statistics\\[8pt]}

\author{%
  \begin{tabular}{c}
  Arul Rana\textsuperscript{1} \quad
  Agrim Tripathi\textsuperscript{1} \quad
  Shoaib Ahmed Dipu\textsuperscript{2} \\[3pt]
  Md. Shaown Miah\textsuperscript{3} \quad
  Syed Ishtiaque Ahmed\textsuperscript{4} \quad
  Sayeed Shafayet Chowdhury\textsuperscript{2} \\[5pt]
  {\normalfont\small
  \textsuperscript{1}Indian Institute of Technology Kharagpur \quad
  \textsuperscript{2}Indiana University Indianapolis} \\[2pt]
  {\normalfont\small
  \textsuperscript{3}Bangladesh University of Engineering and Technology \quad
  \textsuperscript{4}University of Toronto}
  \end{tabular}
  }

\iclrfinalcopy 
\begin{document}
\renewcommand{\headrulewidth}{0pt}
\raggedbottom

\maketitle

\begin{abstract}
Spiking Neural Networks (SNNs) offer an energy-efficient approach to processing event-camera data, yet out-of-distribution (OOD) detection remains challenging in this setting. Existing OOD detection methods often depend on model outputs or computational components that are unavailable in object detection SNNs or are poorly suited to low-compute deployment. To that effect, we show that the subthreshold membrane potential \(V_{\mathrm{mem}}(t)\) provides a useful internal signal for detecting distribution shifts. Simple per-channel statistics derived from these membrane dynamics enable OOD detection. To evaluate this approach, we introduce Gen1-C, an event-camera corruption benchmark developed upon the Prophesee Gen1 automotive detection dataset, containing six sensor-motivated histogram-level stress tests at five severity levels. We further propose the Multi-Descriptor Deviation (MDD), a corruption-blind method that operates on membrane-potential statistics. At the highest corruption severity, MDD achieves an AUROC of more than 0.88 on five of the six corruptions using only a bounded 64-frame observation window. Notably, the remaining corruption is also the one that has the smallest effect on the underlying detector. These results show that the temporal membrane-potential dynamics can provide an effective and low-cost signal for OOD detection in SNN-based event perception.

\end{abstract}

\section{Introduction}
\label{sec: intro}
Event cameras record per-pixel brightness changes as they happen, with microsecond resolution and high dynamic range. Paired with spiking neural networks (SNNs), which also run at low power, they make an efficient sensing-and-processing pipeline for safety-critical uses (self-driving cars, robotics), where one question matters most: if the camera’s input drifts away from what the network was trained on, can the system tell? Out-of-distribution (OOD) detection is the usual answer for ANNs, but nearly every existing method needs something a low-power spiking network detector does not cheaply produce such as class scores, input gradients or a backward pass, or a separate, non-spiking feature network, none of which a spiking network, which is our main motivation to monitor for this paper, produces on its own. 

We start from an internal state already computed by the spiking front end. Reading this value does not require an additional forward pass through the task
network; the resulting descriptors are computed as a separate post-processing stage. A spiking neuron keeps a sub-threshold membrane potential Vmem(t), an internal value the network already updates at every time step to decide whether to fire. Reading this value does not require an additional forward pass through the task network; the resulting descriptors are computed as a separate post-processing stage. We call this vector $\varphi$ and ask how effective it is for detecting these event-camera corruptions, each designed to isolate a distinct functional axis of sensor degradation.
\newpage
\paragraph{Contributions.}
\begin{itemize}[leftmargin=*,itemsep=2pt,topsep=2pt]
\item \textbf{A membrane OOD signal and the Gen1-C benchmark.} We introduce $\varphi$, a summary of the spiking neuron membrane state that can be used as an OOD signal, and release \textbf{Gen1-C}: a reproducible (seeded, model-agnostic) event-camera corruption benchmark with six  corruptions developed upon the Prophesee Gen1 automotive detection dataset \citep{detournemire2020gen1}, motivated by sensor-degradation mechanisms. We provide Gen1-C to facilitate rigorous, standardized evaluation in future research on event-based robustness.
\item \textbf{The Multi-Descriptor Deviation (MDD).}
MDD uses $\varphi$ to score model inputs along several complementary directions in feature space and combines the resulting scores with an OR rule. It is fit using only clean in-distribution data, without access to corruption samples.
    .
\end{itemize}

\section{Background and Related Work}
\label{rel}
\textbf{OOD detection for ANNs}: Most Out-of-Distribution (OOD) detection methods rely on specific network outputs, feature representations, or auxiliary computations. MSP uses softmax probabilities~\citep{hendrycks2017msp}, while ODIN additionally requires temperature scaling and input perturbations~\citep{liang2018odin}. Other methods depend on internal features, such as Mahalanobis distance~\citep{lee2018mahalanobis}, ReAct~\citep{sun2021react}, and ViM~\citep{wang2022vim}, or require gradients through backpropagation, as in GradNorm~\citep{huang2021gradnorm}. KNN-based methods also require access to stored in-distribution feature representations~\citep{sun2022knn}. MDD reduces this memory burden using PCA and reference set reduction.

\textbf{OOD detection for SNNs}: Existing OOD detection methods for SNNs have mainly focused on image classification. \citet{martinez2023snnood} detect OOD samples using atypical spike-count patterns from hidden layers. \citet{terres2024forwardforward} use sparse latent representations learned with Forward-Forward training and measure the distance to class-representative manifolds. \citet{avramovic2025residual} study OOD image detection with residual SNN architectures.

\textbf{Event-camera robustness and corruption benchmarks}:
Robustness benchmarks such as ImageNet-C established the use of controlled, severity-graded corruptions for evaluating model robustness~\citep{hendrycks2019imagenetc}. For event cameras, N-ImageNet introduced variants recorded under different camera trajectories and lighting conditions to evaluate robustness~\citep{kim2021nimagenet}. The v2e framework further models realistic event-camera effects such as threshold mismatch, limited bandwidth, and intensity-dependent noise~\citep{hu2021v2e}. For automotive event-based detection, we use the Prophesee Gen1 dataset, which contains 39 hours of driving data with bounding-box annotations for cars and pedestrians~\citep{detournemire2020gen1}.

\section{Method}
\label{Method}

\subsection{Corruption Suite}
\def\ONcoords{0.40/0.45, 0.70/0.80, 1.00/1.10, 1.30/1.40, 1.62/1.62, 2.00/0.60}
\def\OFFcoords{0.60/1.25, 1.10/0.60, 1.72/1.05, 2.12/1.50}

\newcommand{\drawON}[1]{%
  \foreach \x/\y in \ONcoords {%
    \fill[#1] (\x,\y) circle (1.5pt);
  }%
}

\newcommand{\drawOFF}[1]{%
  \foreach \x/\y in \OFFcoords {%
    \fill[#1] (\x,\y) circle (1.5pt);
  }%
}

\newcommand{\evtframe}[1]{%
  \draw[rounded corners=2pt,fill=black!3,draw=black!45]
    (0,0) rectangle (2.6,2.0);
  #1%
}

\newcommand{\fxhotpix}{%
  \foreach \x/\y in {0.52/1.80, 2.30/1.78, 1.45/0.28}{%
    \fill[red] (\x,\y) circle (2.2pt);
    \draw[red!55] (\x,\y) circle (3.4pt);
  }%
}

\newcommand{\fxflood}{%
  \fill[red!22] (1.65,1.05) ellipse (0.62 and 0.46);
  \foreach \x/\y in {1.30/1.00,1.55/1.22,1.78/0.92,1.86/1.30,1.50/0.82,1.95/1.12,1.40/1.30}{%
    \fill[red!85] (\x,\y) circle (1.4pt);
  }%
}

\newcommand{\fxratemore}{%
  \foreach \x/\y in {0.30/1.70,0.95/0.35,1.45/1.85,2.25/0.95,0.55/0.95,1.20/1.75,1.85/1.55,2.40/1.30,0.80/1.45,1.55/0.45}{%
    \fill[red!55] (\x,\y) circle (1.3pt);
  }%
}

\newcommand{\fxjitter}{%
  \foreach \x/\y in \ONcoords {%
    \fill[black!35] ({\x+0.16},{\y-0.13}) circle (1.5pt);
  }%
  \foreach \x/\y in \ONcoords {%
    \fill[black!22] ({\x-0.13},{\y+0.10}) circle (1.3pt);
  }%
}

\newcommand{\fxdrop}{%
  \fill[white] (1.55,0.0) rectangle (2.6,2.0);
  \draw[dashed,black!55] (1.55,0.0) rectangle (2.6,2.0);
}

\newcommand{\corrPanelC}[2]{%
  \begin{tikzpicture}
    \node[font=\footnotesize\bfseries,anchor=south west]
      at (0,2.06) {#1};

    \begin{scope}
      \evtframe{\drawON{red!75}\drawOFF{blue!70}}
    \end{scope}

    \node[font=\tiny,anchor=north] at (1.3,-0.04) {normal};

    \draw[-{Latex[length=2mm]},thick,black!70]
      (2.72,1.0) -- (3.42,1.0);

    \node[font=\tiny,anchor=south,black!70]
      at (3.07,1.04) {fault};

    \begin{scope}[shift={(3.55,0)}]
      \evtframe{#2}
    \end{scope}

    \node[font=\tiny,anchor=north] at (4.85,-0.04) {corrupted};
  \end{tikzpicture}%
}

\begin{figure}[H]
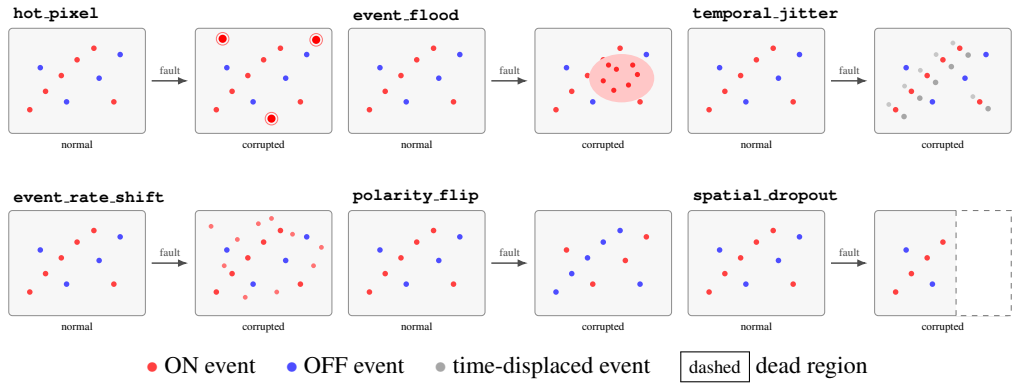

  \centering
  \scriptsize
  \resizebox{0.95\linewidth}{!}{%
    \begin{tabular}{@{}c@{\hskip 3mm}c@{\hskip 3mm}c@{}}
      \corrPanelC{\texttt{hot\_pixel}}{\drawON{red!75}\drawOFF{blue!70}\fxhotpix}
      &
      \corrPanelC{\texttt{event\_flood}}{\drawON{red!75}\drawOFF{blue!70}\fxflood}
      &
      \corrPanelC{\texttt{temporal\_jitter}}{\drawON{red!75}\drawOFF{blue!70}\fxjitter}
      \\[5mm]
      \corrPanelC{\texttt{event\_rate\_shift}}{\drawON{red!75}\drawOFF{blue!70}\fxratemore}
      &
      \corrPanelC{\texttt{polarity\_flip}}{\drawON{blue!70}\drawOFF{red!75}}
      &
      \corrPanelC{\texttt{spatial\_dropout}}{\drawON{red!75}\drawOFF{blue!70}\fxdrop}
    \end{tabular}%
  }

  \vspace{1mm}

  {\footnotesize
    \textcolor{red!75}{$\bullet$}~ON event \quad
    \textcolor{blue!70}{$\bullet$}~OFF event \quad
    \textcolor{black!35}{$\bullet$}~time-displaced event \quad
    \fbox{\scriptsize dashed}~dead region
  }
  \caption{The six Gen1-C corruptions: normal (left) versus corrupted (right) event frame.}
  \label{fig:schematic}
\end{figure}
\textbf{Input Representation} - The main dataset used for experimentation in the paper is the GEN1 Automotive Detection Dataset~\citep{detournemire2020gen1}. We represent each sequence from the dataset by a stack of event-histograms of shape (T, 2, H, W), H x W = 240 x 304. Here, (T = 10) is the number of temporal bins of 5 ms each, and the 2nd dimension records the ON and OFF counts for H x W. Since one frame contains 10 of these bins, it contains information of a 50 ms timeframe.

Each corruption operates on event-histograms and returns a corrupted copy, with a dial of how severe you want the corruption (5 different severities are tested in this paper). We chose transformations whose primary interventions lie along different aspects of the signal. The goal was to minimize overlap between corruption mechanisms. Exact parameter schedules, random seeds, sequence-level application, and corruption geometry checks are given in Appendix~\ref{app:corruptions}.

\begin{itemize} [leftmargin=*,itemsep=2pt,topsep=2pt]
    \item \textbf{Hot Pixels:} We perturb a fixed set $P$ of $n_{\mathrm{hot}}$ pixel locations, chosen randomly per sequence, by adding a constant $\delta$ count to those locations. In real sensors, similar defects arise due to defective circuitry, leakage current, etc.

    \item \textbf{Event Flood:} This corruption injects a burst of extra counts into a small number of localized space-time regions. For $b$ sampled centres, a patch covering a fraction $f$ of the image receives $\delta$ additional counts for a central frame and the two frames immediately before and after it, clipped to the valid range. In practice, this resembles electromagnetic interference, flickering illumination, or a sensor region overwhelmed by saturation.

    \item \textbf{Temporal Jitter:} The On and OFF bin blocks are shifted independently by random integer offsets $s_{\mathrm{ON}}, s_{\mathrm{OFF}} \in [-s_{\max},s_{\max}]$, with bins shifted outside the valid range zeroed out. This disrupts temporal ordering and causes some boundary counts to be lost. It models timestamp noise, clock drift, or readout timing jitter in sensor circuitry.

    \item \textbf{Event-rate Shift:} A single sequence-level scale factor $a$ is applied to the global magnitude of activity, with $a=u<1$ for an under-rate condition and $a>1$ for an over-rate condition. The whole sequence's event density shifts uniformly. This corresponds to systematic changes in event rate caused by sensor bias, scene texture, etc.

    \item \textbf{Polarity Flip:} This corruption targets the sign of events while leaving their count, location, and timing untouched. For a randomly selected fraction $p$ of frames, the On and OFF blocks are swapped.

    \item \textbf{Spatial Dropout:} Rather than adding or shifting activity, this corruption removes it entirely from a region. $K$ rectangular regions are sampled, and all bins and polarities within them are zeroed out across the whole sequence. This models dead-pixel clusters, physical sensor damage, or field-of-view blockage.
\end{itemize}

\subsection{Membrane-Potential Descriptors}

The corruption suite perturbs the magnitude, temporal structure, polarity, and spatial support of event activity. We construct two complementary descriptors from membrane potentials from the SNN model: a pooled descriptor, $\varphi$, and a spatial descriptor, $\varphi_{\mathrm{spatial}}$, to detect these corruptions.
Appendix~\ref{app:theory} gives more details.

\subsubsection{Pooled Membrane Statistics: \texorpdfstring{$\varphi$}{phi}}

For each monitored channel, we summarize the membrane responses across each frame (10 bins) using their mean, variance, and excess kurtosis. These statistics are chosen to capture response magnitude, variation across bins, and sensitivity to extreme responses. We then apply global average pooling (GAP) to each moment map, obtaining three scalar features per channel:
\[
\varphi_c =
\left[
\operatorname{GAP}(\mu_c),
\operatorname{GAP}(\nu_c),
\operatorname{GAP}(\kappa_c)
\right],
\qquad
\operatorname{GAP}(f) =
\frac{1}{HW}\sum_z f(z)
\]

GAP produces a compact descriptor but discards the spatial arrangement of the moment maps, motivating a complementary spatial descriptor.

\subsubsection{Spatial Organization and Persistence: \texorpdfstring{$\varphi_{\mathrm{spatial}}$}{phi\_spatial}}

We construct $\varphi_{\mathrm{spatial}}$ from the same statistics, mean and variance, but do not collapse them using GAP. From these, we compute three groups of measurements:

\begin{itemize} [leftmargin=*,itemsep=2pt,topsep=2pt]
    \item \textbf{Spatial dispersion:} For each channel, we measure the spatial variance of its membrane mean map and the participation ratio of its bin-wise variance map. Spatial variance measures how unevenly the mean response is distributed across the map. Participation ratio measures whether the response variation is broadly spread out or concentrated in a small region:
    \[
    P_c =
    \frac{\left(\sum_z v_c(z)\right)^2}
    {D\sum_z v_c(z)^2}
    \]

    \item \textbf{Local Flatness:} This measures anomalies that a whole-map statistic would average away. We compute variation within local windows, relative to each channel's overall spatial variation:
    \[
    r_c(\mathcal{A};m)
    =
    \frac{\text{local std. dev. within } \mathcal{A}}
    {\text{channel's overall spatial std. dev.}}
    \]
    
    We summarize this at multiple window scales, keeping the most extreme value at each scale. We repeat this using smaller $3\times3$ windows, and from the resulting map extract several low quantiles and the fraction of the map falling below a few fixed thresholds.

    \item \textbf{Persistence:} A flat region might arise naturally within a single frame. To distinguish this from a defect, we track how consistently each location remains flat across successive frames. We summarize the resulting persistence map using upper quantiles and threshold exceedance fractions.
\end{itemize}

\subsection{Multi-Descriptor Deviation (MDD)}

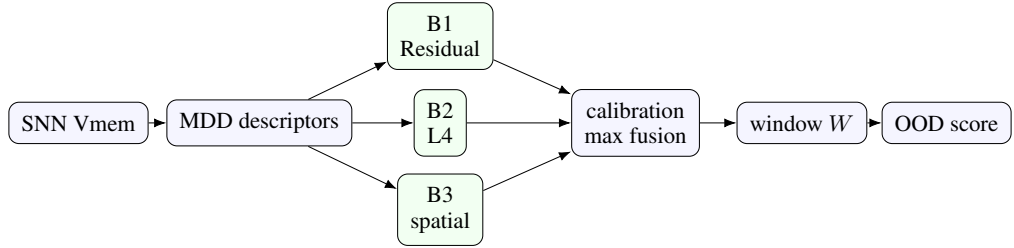
\begin{figure}[H]
  \centering
  \resizebox{0.95\linewidth}{!}{%
  \begin{tikzpicture}[font=\small,>=Latex,
    box/.style={draw,rounded corners,align=center,fill=blue!4,inner sep=5pt},
    br/.style={draw,rounded corners,align=center,fill=green!5,inner sep=5pt}]
    \node[box] (vmem) at (0,0.8) {SNN Vmem};
    \node[box] (desc) at (2.5,0.8) {MDD descriptors};
    \node[br] (b1) at (5.0,2.0) {B1\\Residual};
    \node[br] (b2) at (5.0,0.8) {B2\\L4};
    \node[br] (b3) at (5.0,-0.4) {B3\\spatial};
    \node[box] (fuse) at (7.7,0.8) {calibration\\max fusion};
    \node[box] (window) at (10.0,0.8) {window $W$};
    \node[box] (out) at (12.0,0.8) {OOD score};
    
    \draw[->] (vmem)--(desc);
    \draw[->] (desc)--(b1);
    \draw[->] (desc)--(b2);
    \draw[->] (desc)--(b3);
    
    \foreach \b in {b1,b2,b3}{\draw[->] (\b)--(fuse);}
    
    \draw[->] (fuse)--(window);
    \draw[->] (window)--(out);
  \end{tikzpicture}}
  \caption{The Multi-Descriptor Deviation Detector.}
  \label{fig:mdd-flow-current}
\end{figure}

Corruptions can alter magnitude, channel relationships, spatial organization, and other properties of the inputs. Hence, we created a detector which contains 3 complementary branches which together form MDD. MDD is corruption free, or more simply, it has not seen any corrupted data or information during training/calibration. This is because in application we may find more types of corruption we have not covered in our suite.

The complete clean-only fitting, calibration, normalization, and fusion procedure is specified in Appendix~\ref{app:updated-interface}.

\subsubsection{PCA Residual (B1: Residual)}

We first standardize the descriptor $\varphi$ using the clean-fit feature means and scales. PCA is then fitted on these clean descriptors, and the first $k$ principal directions are retained. Instead of measuring distance within this dominant clean-data subspace,
we measure the energy discarded by PCA:
\[
a_1 =
\left[
\|\tilde{\varphi}-\mu_{\mathrm{PCA}}\|_2^2
-
\|z\|_2^2
\right]^{1/2}.
\]

This branch is intended to detect descriptor changes in low-variance directions that PCA treats as unimportant. In other words, it responds when the membrane-potential descriptor contains structure that lies outside the normal clean-data subspace, complementing the deep-layer
and spatial branches.

\subsubsection{Deep-Layer Deviation (B2: L4)}

Pooling information across layers can dilute deviations that are concentrated in a single layer. We therefore model the deepest-layer descriptor $h$ separately using a Ledoit–Wolf covariance estimate~\citep{ledoit2004well}, yielding the Mahalanobis distance $d_{L4}^2$. We use its absolute deviation from the clean calibration median, normalized by the calibration MAD, so that both unusually high and unusually low distances are anomalous:
\[
a_2=
\frac{\left|d_{L4}^2-\operatorname{median}_{\mathrm{cal}}(d_{L4}^2)\right|}
{\operatorname{MAD}_{\mathrm{cal}}(d_{L4}^2)+\epsilon},
\qquad
d_{L4}^2=(h-\mu_{L4})^\top\Sigma_{L4}^{-1}(h-\mu_{L4}).
\]
\subsubsection{Spatial Dispersion and Organization (B3: Spatial)}

For \(\varphi_{\mathrm{spatial}}\), MDD combines a two-sided
For \(\varphi_{\mathrm{spatial}}\), we clean-fit-standardize the spatial
  dispersion features, compute their Ledoit--Wolf Mahalanobis distance, and use
  its two-sided deviation from the clean-calibration distribution as
  \(q_{\mathrm{disp}}\). To retain local organization, MDD combines four
  layerwise persistence-fraction components with one signed flatness component:
  \[
  c_{\mathrm{flat}}
  =
  \max_{\ell,j}
  s_j\frac{f_{\ell j}-\mu_{\ell j,\mathrm{fit}}}
  {\sigma_{\ell j,\mathrm{fit}}+\epsilon},
  \qquad
  q_{\mathrm{org}}=\max_jF_j^{\mathrm{mid}}(c_j),
  \]
  where \(g_{\ell1},g_{\ell2}\) are nested persistence fractions,
  \(s_j=-1\) for \(j\leq10\) and \(s_j=+1\) otherwise, and
  \(F_j^{\mathrm{mid}}\) is the clean-fit mid-rank percentile of component
  \(c_j\). The deployed spatial score is
  \[
  b_{\mathrm{spatial}}
  =
  \max\left\{
  \frac{q_{\mathrm{disp}}-\mu_{\mathrm{disp,cal}}}
  {\sigma_{\mathrm{disp,cal}}+\epsilon},
  \frac{q_{\mathrm{org}}-\mu_{\mathrm{org,cal}}}
  {\sigma_{\mathrm{org,cal}}+\epsilon}
  \right\}.
  \]
  Thus, either atypical spatial dispersion or anomalous spatial organization
  can activate this branch. Appendix~\ref{app:updated-interface} gives the
  exact feature ordering and clean-reference percentile construction. The
  membrane-dynamics motivation for these statistics and the effect of resetting
  and temporal pooling are analyzed in
  Appendices~\ref{app:reset-error}--\ref{app:linear-response}.

\section{Experimental Results}
\label{results}

\subsection{Motivation - A DETECTOR DEGRADES UNDER CORRUPTIONS}
\label{sec:motivation}
\begin{figure}[H]
\centering
\begin{minipage}[t]{.49\linewidth}
  \centering
  \includegraphics[width=\linewidth]{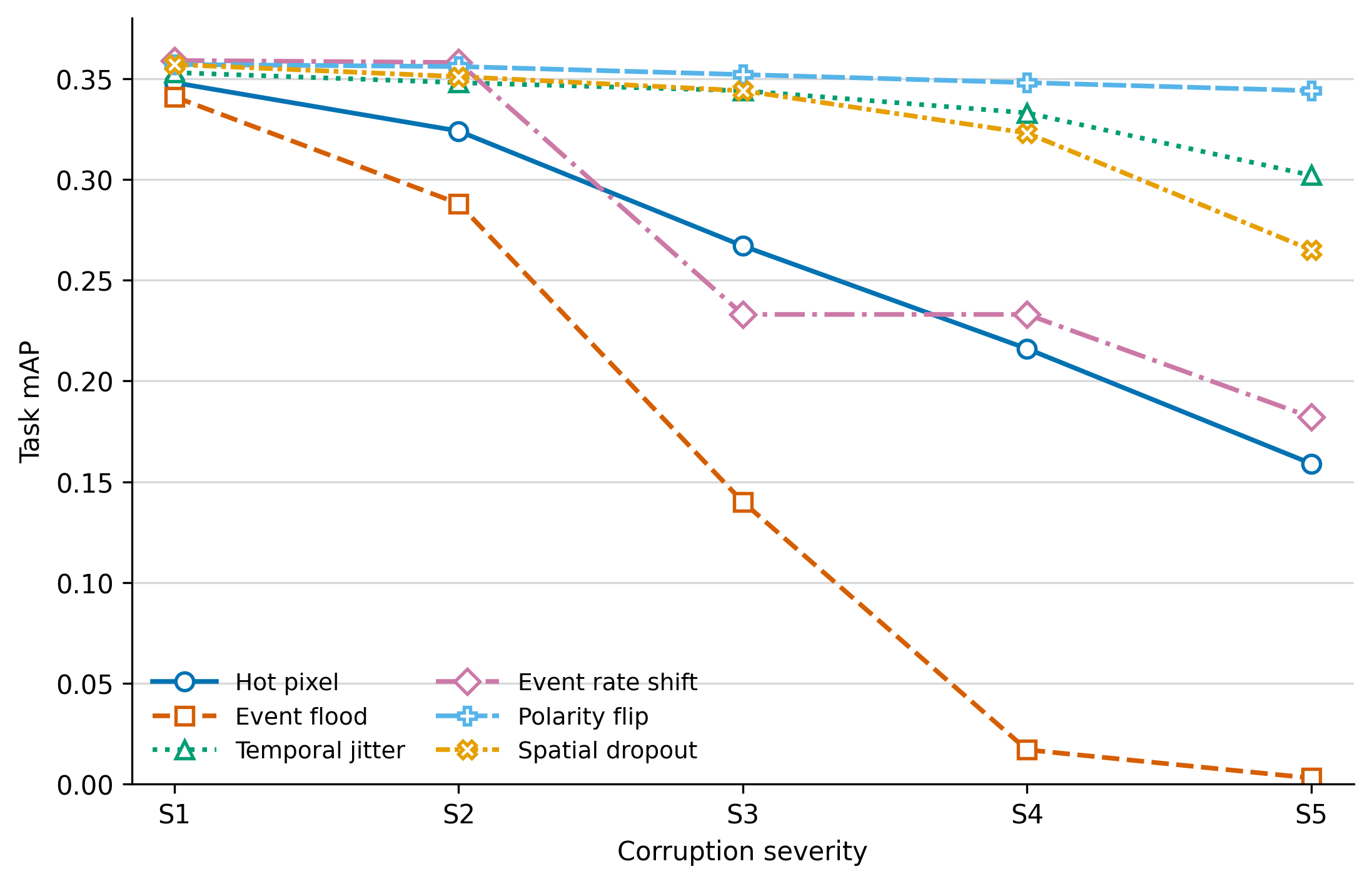}
\end{minipage}\hfill
\begin{minipage}[t]{.49\linewidth}
  \centering
  \includegraphics[width=\linewidth]{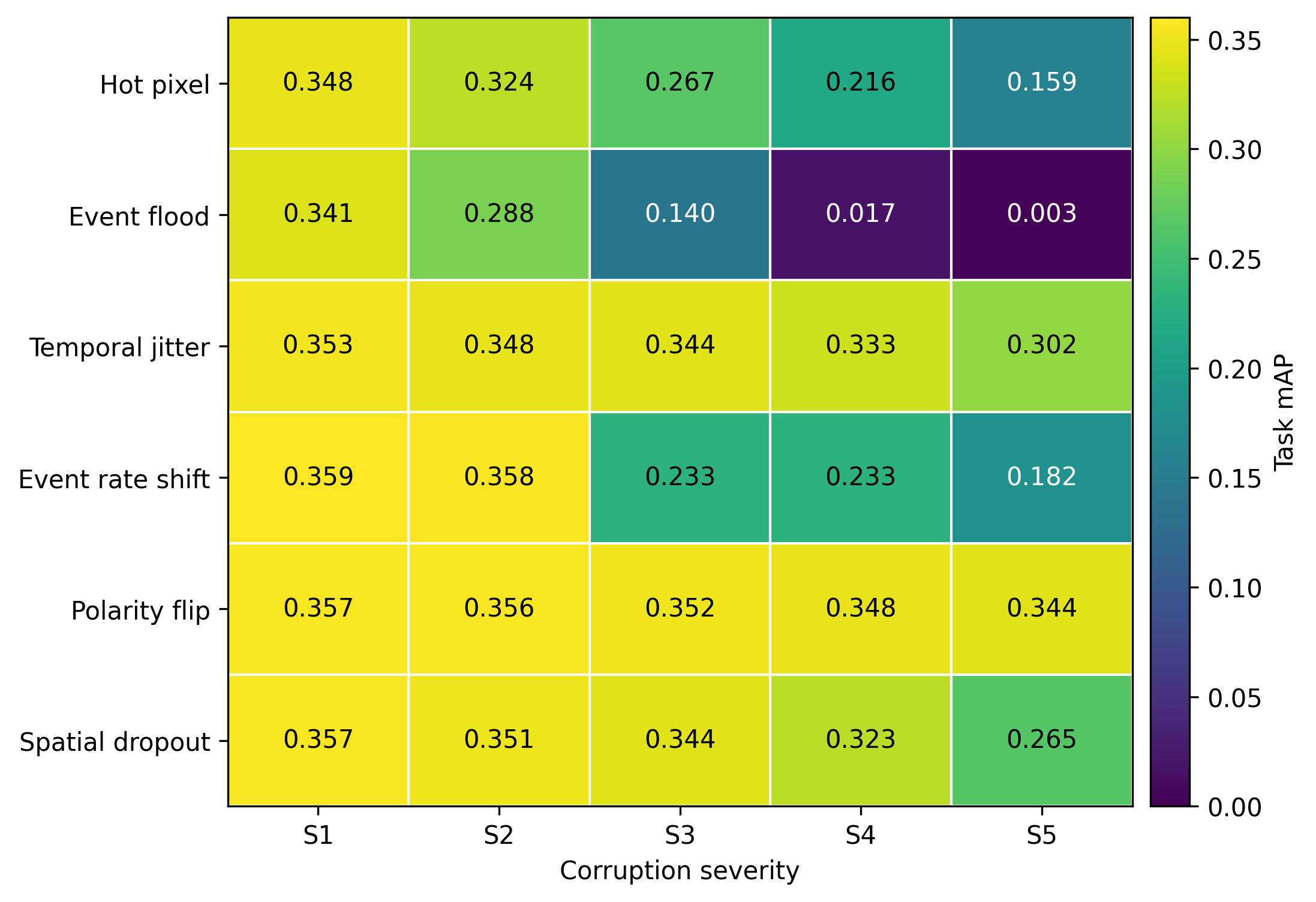}
\end{minipage}
\caption{Task degradation across corruption families and severities.}
\label{fig:corruption-task-degradation}
\end{figure}

We feed the pretrained hybrid SNN--ANN detector~\citep{ahmed2025hybrid} (clean
Gen1 test mAP $\approx0.36$, Fig~\ref{fig:corruption-task-degradation}) each Gen1-C corruption with no retraining, corrupting the event histogram before it is fed into the model. The resulting performance degradation varies substantially across corruption types and severity levels, indicating that the detector is differentially sensitive to sensor-level perturbations. \texttt{event\_flood} is the most destructive (mAP $0.341\to0.003$), with performance deteriorating sharply as severity increases. In contrast, \texttt{polarity\_flip} is nearly harmless (approximately $-4\%$ at S5), showing that not all event-level perturbations substantially affect downstream detection.

\subsection{MDD Setup}
\label{sec:MDDSetup}

We use 302 Gen1 sequences ($\approx$343,099 frames), split by sequence into fitting (50\%), calibration (10\%), sensitivity (10\%), and held-out validation (30\%) sets. Each Gen1-C corruption is applied to a complete sequence at S1--S5. MDD is fit on clean fitting frames only; clean calibration frames provide score normalization and thresholds. We report hyperparameter and fusion sensitivity analyses on the separate sensitivity split in Appendix~\ref{app:mdd-sensitivity}, but do not use its corruption-labelled results to select the final detector, preserving the corruption-blind protocol.

MDD monitors $T=10$ histogram bins from four PLIF layers with $64/128/256/256$ channels ($704$ total). Per-channel $[\mu,\sigma^2,\kappa]$ moments form the $2112$-D pooled descriptor $\varphi$; the spatial descriptor is $1488$-D, comprising $1408$ dispersion features and four $20$-entry organization blocks. MDD combines PCA residual (B1, width $64$), deep-layer deviation (B2), and spatial dispersion/organization (B3) scores by calibrated maximum fusion. Appendix~\ref{app:updated-interface} specifies the descriptor ordering and exact clean-fit/calibration procedure.

Our primary host is hybrid, with a spiking front end followed by an ANN detection pathway. Although ANN features and outputs are available, MDD uses only front-end membrane states, making it independent of the downstream detection head and its output representation. Appendix~\ref{app:generalization} evaluates the same clean-fit procedure on a second, fully spiking detector.

\subsection{Main result: MDD OOD detection}
\label{sec:main}
Table~\ref{tab:mdd} reports fused MDD AUROC per corruption/severity,
per-frame~(a) and per-$64$-window~(b, $W{=}64$, ${\approx}3.2$\,s). One score is already strong per frame on four of the six corruptions at severity~5;
aggregating over the bounded $64$-frame window (Section~\ref{sec:window}) further increases the AUROC for all corruptions. Even on $W{=}64$, \texttt{polarity\_flip} detection struggles, but as we discussed earlier in Section~\ref{sec:motivation}, it barely harms the network even at the highest severity.

\begin{table}[H]
      \caption{MDD AUROC}
      \label{tab:mdd}\label{tab:mdd-frame}\label{tab:mdd-w64}
      \begin{minipage}[t]{0.49\linewidth}\centering\small
      {\footnotesize\textbf{(a) per-frame}}\\[2pt]
      \resizebox{\linewidth}{1.3cm}{%
      \begin{tabular}{lccccc}
        \toprule
        Corruption & S1 & S2 & S3 & S4 & S5 \\
        \midrule
        \texttt{hot\_pixel}         & 1.000 & 1.000 & 1.000 & 1.000 & 1.000 \\
        \texttt{temporal\_jitter}   & 0.673 & 0.724 & 0.775 & 0.825 & 0.863 \\
        \texttt{event\_rate\_shift} & 0.515 & 0.568 & 0.863 & 0.882 & 0.931 \\
        \texttt{spatial\_dropout}   & 0.780 & 0.917 & 0.928 & 0.954 & 0.963 \\
        \texttt{event\_flood}       & 0.504 & 0.508 & 0.517 & 0.531 & 0.551 \\
        \texttt{polarity\_flip}     & 0.505 & 0.509 & 0.518 & 0.532 & 0.545 \\
        \bottomrule
      \end{tabular}}
      \end{minipage}\hfill
      \begin{minipage}[t]{0.49\linewidth}\centering\small
      {\footnotesize\textbf{(b) per-$W{=}64$ window (${\approx}3.2$\,s)}}\\[2pt]
      \resizebox{\linewidth}{1.3cm}{%
      \begin{tabular}{lccccc}
        \toprule
        Corruption & S1 & S2 & S3 & S4 & S5 \\
        \midrule
        \texttt{hot\_pixel}         & 1.000 & 1.000 & 1.000 & 1.000 & 1.000 \\
        \texttt{temporal\_jitter}   & 0.692 & 0.745 & 0.798 & 0.847 & 0.886 \\
        \texttt{event\_rate\_shift} & 0.519 & 0.579 & 0.888 & 0.906 & 0.955 \\
        \texttt{event\_flood}       & 0.556 & 0.602 & 0.692 & 0.795 & 0.892 \\
        \texttt{spatial\_dropout}   & 0.800 & 0.952 & 0.962 & 0.984 & 0.987 \\
        \texttt{polarity\_flip}     & 0.511 & 0.520 & 0.534 & 0.553 & 0.568 \\
        \bottomrule
      \end{tabular}}
      \end{minipage}
  \end{table}

Branch-wise and representation ablations are reported in Section~\ref{sec:mdd-ablation-results}; PCA hyperparameters, fusion method analyses and other experiments are reported in Appendix~\ref{app:mdd-sensitivity}. Cross-dataset transfer to Prophesee Gen4 and cross-model evaluation with a different spiking YOLO detector are reported in Appendix~\ref{app:generalization}. These results show that MDD effectiveness depends not only on the corruption itself, but also on input preprocessing, the representations learned by the host model, and the broader sensor model pipeline.

\subsection{Branch Contributions}
\label{sec:mdd-ablation-results}

All branch ablations below use the S5 validation pool and non-overlapping $W=64$ windows. The branch parameters, clean references, and calibration scales are held fixed while the reported branch or representation is changed.

\begin{table}[t]
\caption{MDD branch contributions and leave-one-out at S5 and $W=64$}
\label{tab:mdd-branch-pair}
\begin{minipage}[t]{0.49\linewidth}
\centering\small
{\footnotesize\textbf{(a) Branch contributions}}\\[2pt]
\resizebox{\linewidth}{1.4cm}{%
\begin{tabular}{lcccc}
\toprule
Corruption & B1 & B2 & B3 \\
\midrule
\texttt{hot\_pixel} & 1.000 & 0.998 & 1.000 \\
\texttt{temporal\_jitter} & 0.805 & 0.851 & 0.648 \\
\texttt{event\_rate\_shift} & 0.507 & 0.763 & 0.938 \\
\texttt{spatial\_dropout} & 0.455 & 0.473 & 0.992 \\
\texttt{event\_flood} & 0.891 & 0.787 & 0.892 \\
\texttt{polarity\_flip} & 0.583 & 0.512 & 0.558 \\
\bottomrule
\end{tabular}}
\end{minipage}\hfill
\begin{minipage}[t]{0.49\linewidth}
\centering\small
{\footnotesize\textbf{(b) Leave-one-branch-out}}\\[2pt]
\resizebox{\linewidth}{1.4cm}{%
\begin{tabular}{lcccc}
\toprule
Corruption & Full & $-$B1 & $-$B2 & $-$B3 \\
\midrule
\texttt{hot\_pixel} & 1.000 & 1.000 & 1.000 & 1.000 \\
\texttt{temporal\_jitter} & 0.886 & 0.846 & 0.887 & 0.830 \\
\texttt{event\_rate\_shift} & 0.955 & 0.938 & 0.955 & 0.745 \\
\texttt{spatial\_dropout} & 0.987 & 0.991 & 0.987 & 0.478 \\
\texttt{event\_flood} & 0.892 & 0.893 & 0.892 & 0.892 \\
\texttt{polarity\_flip} & 0.568 & 0.558 & 0.568 & 0.598 \\
\bottomrule
\end{tabular}}
\end{minipage}
\end{table}

The branches respond to different shift types: the deep-layer branch is strongest individually for temporal jitter, while the spatial branch dominates spatial dropout and event-rate shift. The PCA-residual B1 branch is weak in isolation but adds complementary information to the fused detector. Leave-one-out results show that removing B3 collapses dropout detection ($0.987\to0.478$), whereas removing B2 has little effect on fused jitter performance ($0.886\to0.887$); removing B1 reduces jitter from $0.886$ to $0.846$.

\subsection{Representation Ablation}

We compare the representations using complementary distributional
statistics. Jensen--Shannon divergence measures the difference between the clean and corrupted score distributions. MMD$^2$ compares their raw vector distributions using a kernel-based two-sample statistic, without fitting a detector~\citep{gretton2012kernel}. C2ST balanced accuracy measures how well a held-out binary classifier distinguishes clean from corrupted samples~\citep{lopezpaz2017revisiting}. AUPR and FPR@95 summarize ranking quality and operating-point behavior.

\begin{table}[H]
\centering
\caption{Distributional diagnostics.}
\label{tab:representation-diagnostics}

\begin{minipage}{\linewidth}
\centering\small
\scalebox{0.85}{%
\begin{tabular}{lrrrrr}
\toprule
Representation & AUPR & FPR@95 & JS bits & MMD$^2$ & C2ST BA \\
\midrule
Full membrane $\varphi$ & 0.637 & 0.818 & 0.158 & 0.053 & 0.695 \\
$\mu$ & 0.602 & 0.810 & 0.148 & 0.075 & 0.693 \\
$\sigma^2$ & 0.569 & 0.944 & 0.025 & 0.027 & 0.666 \\
$\kappa$ & 0.589 & 0.880 & 0.009 & 0.016 & 0.653 \\
$\varphi_{\mathrm{spatial}}$ & 0.640 & 0.786 & 0.184 & 0.173 & 0.750 \\
$\varphi+\varphi_{\mathrm{spatial}}$ & 0.650 & 0.821 & 0.159 & 0.147 & 0.771 \\
\bottomrule
\end{tabular}%
}
\end{minipage}
\end{table}

The distributional diagnostics reveal complementary behavior across the representations. The spatial representation has the largest average Jensen--Shannon divergence ($0.184$) and MMD$^2$ ($0.173$) under the reported diagnostic settings. It is particularly effective for hot pixels and event-rate shifts. The concatenated representation provides the strongest supervised separation, with the highest C2ST balanced accuracy ($0.771$) and AUPR ($0.650$), showing that temporal and spatial cues jointly provide useful information
to a flexible classifier. In contrast, polarity flip shows no robust separation: its JS divergence is near zero across representations, the raw-vector MMD$^2$ tests are non-significant, and C2ST balanced accuracy remains close to chance. Thus, the small polarity-flip differences are not sufficiently consistent across recordings to support a reliable detector.

\subsection{Calibration and Uncertainty}
\label{sec:mdd-calibration-results}

\begin{table}[H]
\centering\small
\caption{MDD operating points at S5, $W=64$.}
\label{tab:mdd-calibration-results}
\begin{tabular}{lrrrrrrr}
\toprule
Calibration percentile & Clean FPR & Hot pixel & Jitter & Rate shift & Flood & Dropout & Polarity flip \\
\midrule
85\% & 0.146 & 1.000 & 0.713 & 0.893 & 0.812 & 1.000 & 0.216 \\
90\% & 0.099 & 1.000 & 0.630 & 0.879 & 0.803 & 1.000 & 0.168 \\
95\% & 0.028 & 1.000 & 0.551 & 0.851 & 0.788 & 1.000 & 0.092 \\
99\% & 0.005 & 1.000 & 0.389 & 0.349 & 0.783 & 0.008 & 0.034 \\
\bottomrule
\end{tabular}
\end{table}

The operating points expose the ranking--alarm trade-off. Hot pixels remain detectable at every threshold, while dropout is fully detected through the 95th percentile at a held-out clean FPR of $2.8\%$; its TPR collapses at the
99th percentile because the extreme threshold excludes most affected windows. Rate shift and flood retain useful recall at the 95th percentile ($85.1\%$ and $78.8\%$, respectively), whereas jitter decreases more steadily.
Polarity flip remains weak at every operating point, reaching only $9.2\%$ TPR at the 95th-percentile threshold.

\subsection{Windowed aggregation: the deployable operating point}
\label{sec:window}

\begin{figure}[H]
  \centering
  \includegraphics[width=0.72\linewidth]{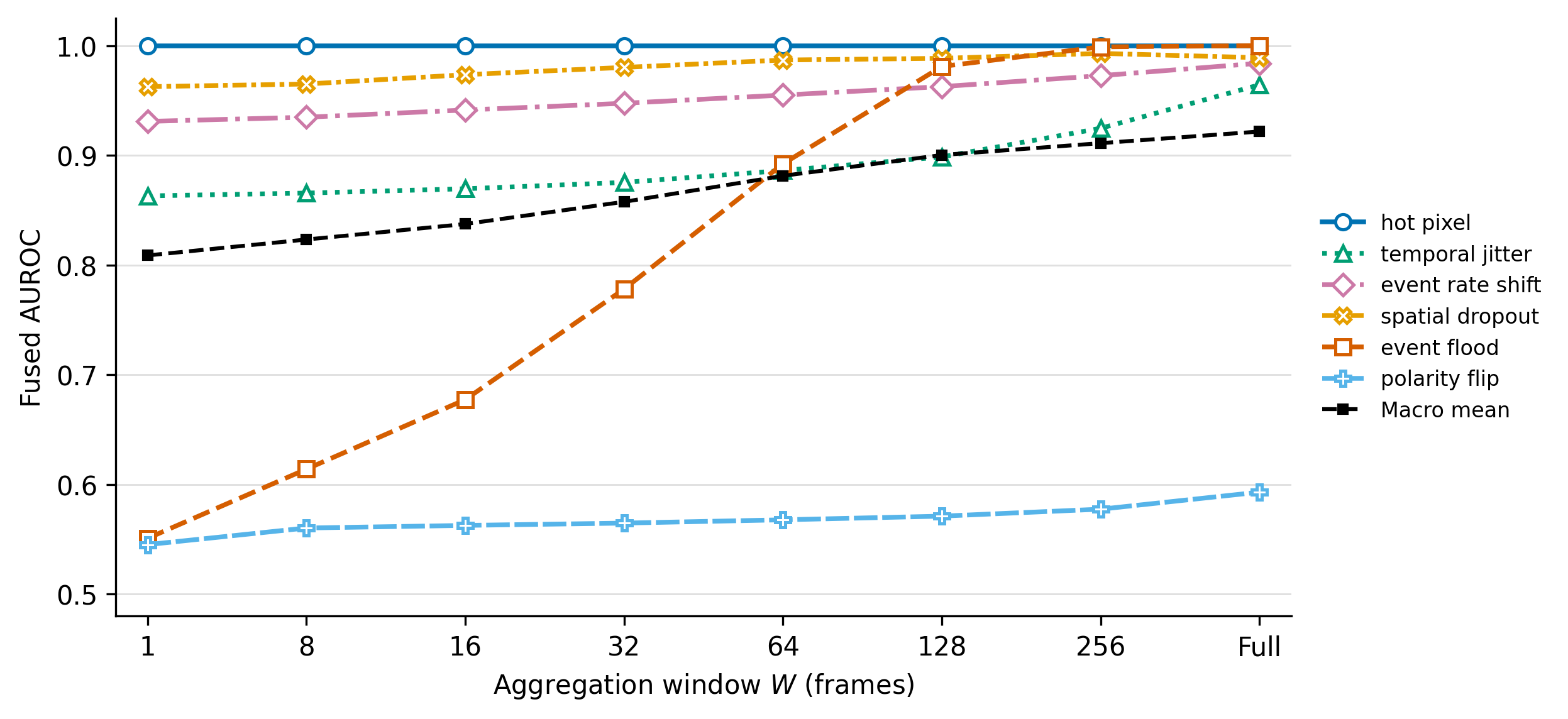}
  \caption{Fused AUROC vs.\ aggregation window $W$ (S5)}
  \label{fig:window}
\end{figure}

Table~\ref{tab:mdd-w64} reports results using $W{=}64$ consecutive frames, corresponding to approximately $3.2$\,s per decision. Each window score is the average of its $W$ frame-level MDD scores. Since each input frame spans $50$\,ms for this model, the observation duration is $T_{\mathrm{obs}} = W \times 0.05$\,s. This operating point targets sustained degradation or drift rather than acute failures requiring sub-second alarms. Figure~\ref{fig:window} shows the severity-5 AUROC sweep over window lengths.

\paragraph{Windowing versus per-frame signal.}
For \texttt{event\_flood}, windowing integrates a temporally consistent bias, producing the large $0.55\!\to\!0.98$ gain. \texttt{polarity\_flip} remains close to chance ($0.55\!\to\!0.57$ through $W{=}128$). In
contrast, \texttt{spatial\_dropout} already has a substantial per-frame signal ($0.96$) and improves modestly to $0.99$ with windowing.

The full window-length sweep is reported in Appendix~\ref{app:mdd-sensitivity}.

\subsection{Matched OOD-detector benchmark}
\label{sec:mdd-baseline-benchmark}

We compare MDD with the compatible post-hoc detectors in our matched baseline suite under the same Gen1-C sequences, corruption severities and recording split.  We adapt these feature-based baselines to the monitored SNN representations.  The suite includes deep nearest neighbors~\citep{sun2022knn}, DIME \citep{sjogren2021dime}, PatchCore~\citep{roth2022patchcore}, a Gaussian Mahalanobis control~\citep{lee2018mahalanobis}, and a feature-density flow baseline~\citep{cook2024featuredensity}.

\begin{table}[H]
\centering\scriptsize
\caption{Matched OOD-detector benchmark at S5.}
\label{tab:mdd-baseline-benchmark}
\setlength{\tabcolsep}{3pt}
\resizebox{0.85\linewidth}{!}{%
\begin{tabular}{lccccccc}
\toprule
Method & Hot pixel & Jitter & Rate shift & Dropout & Flood & Polarity & Mean\\
\midrule
MDD (ours) & \textbf{1.000} & \textbf{0.863} & \textbf{0.931} & \textbf{0.963} & 0.551 & \textbf{0.545} & \textbf{0.809}\\
Membrane-kNN & 0.930 & 0.589 & 0.380 & 0.499 & 0.539 & 0.493 & 0.572\\
Spike-count-kNN & 0.973 & 0.400 & 0.341 & 0.451 & 0.518 & 0.491 & 0.529\\
DIME & 0.853 & 0.541 & 0.703 & 0.460 & 0.511 & 0.484 & 0.592\\
Gaussian Mahalanobis & 0.909 & 0.550 & 0.697 & 0.464 & 0.514 & 0.487 & 0.603\\
PatchCore & 0.067 & 0.457 & 0.660 & 0.509 & 0.535 & 0.525 & 0.459\\
FDE-Glow adaptation & 0.974 & 0.605 & 0.665 & 0.464 & 0.512 & 0.524 & 0.624\\
\bottomrule
\end{tabular}}
\end{table}

The matched baselines use the same clean fitting data, calibration split and temporal history as MDD.

\subsection{Efficient MDD for Resource-Constrained Hardware}
\label{sec:efficient-mdd}

To complement the low-compute SNN, we implement Efficient MDD: an MCU-oriented streaming float32 C version of MDD operating directly on sampled post-step membrane potentials. It preserves clean-fit multi-branch deviation
scoring while using fixed-cost streaming computations. It samples four fixed channels from each PLIF layer at strides $(8,4,4,4)$, shares temporal and spatial statistics across branches, evaluates four compact 20-dimensional layerwise Mahalanobis scores and one recurrent flatness score, and fuses the clean-calibrated scores by their maximum. PCA residuals, rank lookup, spatial sorting, and other expensive operations are removed. Appendix~\ref{app:efficient-mdd-app} gives the descriptor, state, implementation, and validation details.

At severity~5, Efficient MDD obtains macro AUROC $0.7694$ per frame and $0.8309$ with $W=64$ windows. The original version obtains $0.8090$ and $0.8813$, respectively. Windowing averages approximately $3.2$\,s of observations and does not change the measured per-frame compute time.

\begin{table}[H]
\centering
\caption{Efficient MDD results. \textbf{(a)} Measured latency per frame \textbf{(b)} Severity-5 AUROC with $W=64$.}
\label{tab:efficient-mdd-results}

\begin{minipage}[t]{0.50\linewidth}
\centering\small
{\footnotesize\textbf{(a) Native-C latency}}\\[2pt]
{\renewcommand{\arraystretch}{1.15} 
\resizebox{\linewidth}{!}{%
\begin{tabular}{lccc}
\toprule
Target & Captured max (ms) & $<50$~ms\\
\midrule
STM32F103RE (M3)      & 247.05 & No\\
STM32L475 (M4F)        & 10.84 & Yes\\
nRF52840 (M4F)         & 13.62 & Yes\\
nRF52832 DK (M4F)      & 19.42 & Yes\\
DWM3001 nRF52833 (M4F) & 19.46 & Yes\\
\bottomrule
\end{tabular}}}
\end{minipage}
\hfill
\begin{minipage}[t]{0.46\linewidth}
\centering\small
{\footnotesize\textbf{(b) Windowed AUROC, $W=64$}}\\[2pt]
{\setlength{\tabcolsep}{3pt}\renewcommand{\arraystretch}{0.85}%
\begin{tabular}{@{}lcc@{}}
\toprule
Corruption & Reference & Efficient\\
\midrule
\texttt{hot\_pixel}         & 1.000 & 1.000\\
\texttt{event\_flood}       & 0.892 & 0.888\\
\texttt{temporal\_jitter}   & 0.886 & 0.748\\
\texttt{event\_rate\_shift} & 0.954 & 0.888\\
\texttt{spatial\_dropout}   & 0.986 & 0.926\\
\texttt{polarity\_flip}     & 0.567 & 0.534\\
\midrule
Macro                       & 0.881 & 0.830\\
\bottomrule
\end{tabular}}
\end{minipage}
\end{table}

The latency measurements cover the native C monitor from sampled membrane potentials to final score, excluding SNN execution, raw-Vmem gathering, transfer etc. The model produces one input frame every $50$\,ms; Efficient MDD must therefore
complete scoring within $50$\,ms to sustain real-time operation without backlog. We measure the sampled-Vmem-to-score C path on physical boards in the FIT IoT-LAB remote testbed~\citep{adjih2015iotlab}. All tested M4F targets
remain within this budget, whereas the STM32F103RE (M3, software floating point)has a captured maximum of 247.05\,ms and cannot keep pace with the input stream. The reduced descriptor, fixed scorer, memory footprint, input-traffic accounting, and per-severity results are detailed in Appendix~\ref{app:efficient-mdd-app}.

\section{Limitations and future work.}
MDD almost completely fails on \texttt{polarity\_flip}: its AUROC remains near chance across severities, indicating that the current membrane representation does not retain a useful polarity-sensitive signal. Detection at low severities is also insufficient for all corruption families and requires improved representations, calibration, or temporal evidence. Our evaluation applies one synthetic corruption family at a time; we do not test mixed corruptions, corruption orderings, or interactions between faults. Another limitation is that model-specific data preprocessing can attenuate or discard some raw corruption signals before MDD observes membrane activity. The failiure of Efficient MDD to pass the latency threshold on a lower end board is also an area of improvement.
\newline
Future work could extend MDD in four directions. A severity-aware formulation could estimate corruption intensity in addition to detecting its presence. Compositional evaluation should test mixtures, orderings, and changing severities of faults rather than isolated interventions. Branch disagreement could also provide an uncertainty or abstention signal for cases outside the calibration distribution. Finally, the spatial and layer-specific branches could be used for fault localization, identifying which layers or image regions contribute most strongly to an alarm.

\section{Conclusion}
MDD turns intermediate membrane activity into a practical sensor-health signal: it combines signals and then produces one calibrated score. We also introduced a controlled six-family corruption suite allowing sensor changes to be evaluated independently of the primary detection task. Across this suite, the results show strong separation for several faults, while also exposing cases where the representation contains little useful evidence. This work also provides a foundation for future research on robust event-camera monitoring. We created Gen1-C to facilitate rigorous, standardized evaluation in future research on event-based robustness.

\newpage

\bibliographystyle{arxiv}
\bibliography{arxiv}

\newpage

\appendix

\section{Membrane Dynamics and the MDD Interface}
\label{app:theory}

This appendix specifies the descriptors and three-branch MDD scorer used in
the experiments and provides conditional interpretations of membrane responses
to corruption. The analytical results concern simplified membrane dynamics;
they do not establish monotonicity of the descriptors or the complete detector.

\subsection{Descriptor and Detector Interface}
\label{app:updated-interface}

\paragraph{Pooled descriptor.}
For a monitored channel $c$, let $V_{tc}(z)$ denote the membrane value at
temporal bin $t$ and spatial location $z$. The temporal moment maps are
\begin{equation}
 \nu_c(z)=\max\left\{\frac1T\sum_t(V_{tc}(z)-\mu_c(z))^2,10^{-8}\right\},\qquad
 \kappa_c(z)=\frac{T^{-1}\sum_t(V_{tc}(z)-\mu_c(z))^4}{\nu_c(z)^2}-3.
\end{equation}
The pooled descriptor concatenates, in layer order, the channel vectors of
globally averaged means, variances, and excess kurtoses:
\begin{equation}
 \varphi_c=[\operatorname{GAP}(\mu_c),\operatorname{GAP}(\nu_c),
 \operatorname{GAP}(\kappa_c)],\qquad
 \operatorname{GAP}(m)=\frac1{HW}\sum_z m(z).
 \label{eq:pooled-descriptor}
\end{equation}
The four monitored PLIF layers have $64/128/256/256$ channels, giving
$\varphi\in\mathbb R^{2112}$. Within each layer, all mean entries precede
all variance entries and then all kurtosis entries.

\paragraph{Spatial descriptor.}
The spatial descriptor retains two dispersion statistics per channel and
one organization block per layer:
\begin{equation}
 \varphi_{\rm spatial}=[b_{\rm disp};o_1;o_2;o_3;o_4],\qquad
 b_{\rm disp}\in\mathbb R^{1408},\quad o_\ell\in\mathbb R^{20}.
 \label{eq:spatial-dimension}
\end{equation}
The dispersion statistics are the population spatial variance of $\mu_c$
and the participation ratio of $\nu_c$,
\begin{equation}
 P_c=\frac{(\sum_z\nu_c(z))^2}{D\sum_z\nu_c(z)^2},\qquad D=HW.
 \label{eq:participation-ratio}
\end{equation}
Within each layer's dispersion block, spatial variances precede participation
ratios. The organization blocks use the valid spatial region, excluding padding.

For either moment map $m_c\in\{\mu_c,\nu_c\}$, define
\begin{equation}
 r_c(\mathcal A;m)=
 \frac{\operatorname{std}_{z\in\mathcal A}m_c(z)}
 {\max\{\operatorname{std}_{z}m_c(z),10^{-8}\}}.
 \label{eq:local-flatness}
\end{equation}
All standard deviations here use population normalization. For window scales
$q\in\{1/4,1/8\}$, the implementation uses window dimensions
$\max\{2,\operatorname{round}(qH)\}$ and
$\max\{2,\operatorname{round}(qW)\}$, with strides equal to one quarter of
each window dimension, rounded down and floored at one. It averages $r_c$
over channels at each window position and retains the minimum over positions.
This produces four entries: two scales for each of the two moment maps.

For $3\times3$ windows with stride one, define the cross-channel map
$C_m(z)=\max_c r_c(\mathcal A_z;m)$. For each moment map, the descriptor
stores its $0.001,0.005,0.02$ spatial quantiles and the fractions of locations
with $C_m<0.02,0.05,0.1$. These contribute another twelve entries.
For frame $n$ within a recording, define
\begin{equation}
 I_n(z)=\mathbf1\{C_{\mu,n}(z)<0.05,\ C_{\nu,n}(z)<0.05\},\qquad
 R_n(z)=\frac1n\sum_{k=1}^{n}I_k(z).
\end{equation}
The final four entries are the $0.995,0.999$ quantiles of $R_n$ and the
fractions of locations with $R_n>0.9$ and $R_n>0.99$. Persistence is causal
and resets at recording boundaries. Its quantiles are stored in the descriptor
but are not used by the MDD organization scorer below.

\paragraph{Clean calibration.}
Let $\mathcal F$ and $\mathcal K$ denote the clean fitting and calibration
frames. For a scalar statistic $q$, the score normalization is explicitly
\begin{equation}
 \mathcal C_q(q)=\frac{q-\operatorname{mean}_{\mathcal K}(q)}
 {\operatorname{std}_{\mathcal K}(q)+\epsilon},\qquad \epsilon=10^{-9}.
 \label{eq:score-calibration}
\end{equation}
These affine normalizations are applied separately to the statistics specified
below. They do not convert the final fused score into a probability.

\paragraph{B1: PCA residual.}
Standardize $\varphi$ using clean-fit coordinate means and standard deviations,
then subtract the clean-fit PCA centering vector:
\begin{equation}
 x=\frac{\varphi-\mu_{\mathcal F}}{\sigma_{\mathcal F}+\epsilon}
       -\mu_{\rm PCA},\qquad
 e=\sqrt{\max\{\|x\|_2^2-\|Ux\|_2^2,0\}},
 \label{eq:pca-residual}
\end{equation}
where the rows of $U\in\mathbb R^{64\times2112}$ are the retained orthonormal
PCA directions. The implementation first normalizes $e$ using its clean-fit
mean and standard deviation, then normalizes that statistic on clean calibration:
\begin{equation}
 q_e=\frac{e-\operatorname{mean}_{\mathcal F}(e)}
 {\operatorname{std}_{\mathcal F}(e)+\epsilon},\qquad
 a_1=\mathcal C_{q_e}(q_e).
 \label{eq:branch-one}
\end{equation}
B1 scores the upper tail of reconstruction residual; it does not use an
absolute radius deviation or a direction-conditioned nearest-neighbor score.

\paragraph{B2: deep-layer deviation.}
Let $h\in\mathbb R^{768}$ be the deepest layer's pooled descriptor. With its
clean-fit mean and Ledoit--Wolf covariance estimate, define
\begin{equation}
 d_{\rm L4}^2=(h-\mu_{\rm L4})^\top\Sigma_{\rm L4}^{-1}(h-\mu_{\rm L4}),
 \qquad q_{\rm L4}=\frac{|d_{\rm L4}^2-m_{\rm L4}|}{\delta_{\rm L4}},
 \qquad a_2=\mathcal C_{q_{\rm L4}}(q_{\rm L4}).
 \label{eq:branch-two}
\end{equation}
Here $m_{\rm L4}$ is the median clean-calibration distance, and
$\delta_{\rm L4}$ is its unscaled median absolute deviation plus $\epsilon$.
The absolute deviation allows both unusually large and unusually small distances
to contribute evidence.

\paragraph{B3: spatial dispersion and organization.}
Standardize $b_{\rm disp}$ coordinate-wise using clean-fit statistics and
fit a Ledoit--Wolf covariance in that standardized space. Define
\begin{equation}
 d_{\rm sp}^2=(\tilde b_{\rm disp}-\mu_{\rm sp})^\top
 \Sigma_{\rm sp}^{-1}(\tilde b_{\rm disp}-\mu_{\rm sp}),\qquad
 q_{\rm sp}=\frac{|d_{\rm sp}^2-m_{\rm sp}|}{\delta_{\rm sp}},
 \label{eq:spatial-distance}
\end{equation}
where $m_{\rm sp}$ and $\delta_{\rm sp}$ are the clean-calibration median
and unscaled median absolute deviation plus $\epsilon$.

Write $o_\ell=(f_{\ell1},\ldots,f_{\ell16},p_{\ell1},p_{\ell2},
g_{\ell1},g_{\ell2})$: the first sixteen entries are the local-variation
statistics defined above, the next two are persistence quantiles, and the
last two are persistence fractions. Let
\begin{equation}
 c_{\rm flat}=\max_{\ell=1,\ldots,4\,;\,j=1,\ldots,16}
 s_j\frac{f_{\ell j}-\mu_{\ell j,\mathcal F}}
 {\sigma_{\ell j,\mathcal F}+\epsilon},\qquad
 s_j=\begin{cases}-1,&j\leq10,\\+1,&j>10.\end{cases}
 \label{eq:flatness-score}
\end{equation}
Thus low window minima or low coherence quantiles, and high flat-location
fractions, increase the flatness component. Construct five organization components,
\begin{equation}
 c=[\min(g_{11},g_{12}),\ldots,\min(g_{41},g_{42}),c_{\rm flat}].
\end{equation}
Because the two persistence thresholds are nested, each minimum equals the
fraction above the stricter threshold. For component $j$, let
$c_{ij}^{\rm ref}$ be its value in clean-fit reference frame $i$, using at
most $100{,}000$ fitting frames. Its mid-rank percentile and the organization score are
\begin{equation}
 F_j^{\rm mid}(c_j)=\frac{
 \#\{i:c_{ij}^{\rm ref}<c_j\}+\tfrac12\#\{i:c_{ij}^{\rm ref}=c_j\}}
 {N_{\rm ref}},\qquad
 q_{\rm org}=\max_{j=1,\ldots,5}F_j^{\rm mid}(c_j).
 \label{eq:organization-score}
\end{equation}
The spatial branch and final fusion are exactly
\begin{equation}
 a_3=\max\{\mathcal C_{q_{\rm sp}}(q_{\rm sp}),
                 \mathcal C_{q_{\rm org}}(q_{\rm org})\},\qquad
 a_{\rm MDD}=\max\{a_1,a_2,a_3\}.
 \label{eq:plain-max-fusion}
\end{equation}
In particular, $c_{\rm flat}$ enters through the organization percentile;
it is not fused as a separate uncalibrated score.

\paragraph{Windowing and data use.}
Confidence-window aggregation averages consecutive frame-level fused scores
within each recording. It is separate from the recording history used for
persistence. The sequence-disjoint partition is $50\%$ fitting, $10\%$
calibration, $10\%$ sensitivity testing, and $30\%$ held-out validation.
Descriptor standardization, PCA, covariance estimates, and organization
references use clean fitting frames; score normalization and thresholds use
clean calibration frames.

\subsection{Membrane Dynamics and Reset Error}
\label{app:reset-error}

For severity $\gamma\geq0$, write
\begin{equation}
 X_t(\gamma)=X_t(0)+\gamma\Delta_t+\varepsilon_t(\gamma),\qquad
 \mathbb E[\varepsilon_t(\gamma)]=0.
 \label{eq:theory-input}
\end{equation}
Consider the idealized stable recurrence with subtractive reset,
\begin{equation}
 V_t=\lambda V_{t-1}+WX_t-\theta s_{t-1},\qquad 0<\lambda<1.
 \label{eq:reset-dynamics}
\end{equation}
This recurrence is an analytical approximation, rather than an exact
specification of every host's PLIF update or reset rule. For matched initial
states, the clean-to-corrupted difference decomposes as
\begin{equation}
 \delta V_t=L_t(\gamma)+R_t,\quad
 L_t(\gamma)=\sum_{k=1}^t\lambda^{t-k}W
 (\gamma\Delta_k+\varepsilon_k(\gamma)),\quad
 R_t=-\theta\sum_{k=1}^t\lambda^{t-k}
 (s_{k-1}^{\rm corr}-s_{k-1}^{\rm clean}).
 \label{eq:linear-response}
\end{equation}
For a scalar neuron with binary spikes and disagreement probabilities
$p_k=\Pr(s_k^{\rm corr}\ne s_k^{\rm clean})\leq p$, the triangle inequality gives
\begin{equation}
 \mathbb E|R_t|\leq\theta\sum_{k=1}^t\lambda^{t-k}p_{k-1}
 \leq\theta p\frac{1-\lambda^t}{1-\lambda}.
 \label{eq:reset-error}
\end{equation}
The same bound applies coordinate-wise to a layer under the corresponding
per-neuron disagreement assumption. It is conservative and may become
uninformative when disagreement is large or leakage is weak. A small
$p/(1-\lambda)$ controls the absolute reset-error bound; neglecting resets
relative to $L_t$ additionally requires that this error be small compared
with the input-driven response.

For excitatory shifts that increase firing, additional resets oppose the
driving shift and can attenuate mean membrane changes. Neither this bound
nor the recurrence implies a universal membrane range of $[0,\theta)$.
Variance, kurtosis, and spatial organization can change even when reset
feedback limits mean drift, motivating descriptors beyond the mean alone.

\subsection{Conditional Linear-Response Results}
\label{app:linear-response}

\paragraph{Mean response.}
With matched initial states and the zero-mean noise assumption above,
\begin{equation}
 \mathbb E[\delta V_t]=\gamma b_t+\mathbb E[R_t],\qquad
 b_t=\sum_{k=1}^t\lambda^{t-k}W\mathbb E[\Delta_k].
 \label{eq:mean-drift}
\end{equation}
A mean input perturbation can therefore affect mean membrane statistics
when its temporally accumulated projection $b_t$ is nonzero. Reset feedback
can oppose this response; nonzero input mean alone does not guarantee a
nonzero membrane drift or monotonic MDD score.

\paragraph{Covariance under temporally white input.}
For the reset-free vector model $V_t=\lambda V_{t-1}+WX_t$, with scalar
leak $\lambda$ and temporally uncorrelated stationary input of covariance
$\Sigma_X$, the stationary covariance is
\begin{equation}
 \Sigma_V=\lambda^2\Sigma_V+W\Sigma_XW^\top
       =\frac{W\Sigma_XW^\top}{1-\lambda^2}.
 \label{eq:white-input-covariance}
\end{equation}
Thus positive-semidefinite input covariance growth implies
positive-semidefinite membrane covariance growth in this model. The expression
requires temporally uncorrelated input; it does not directly describe correlated
event bins or nonlinear reset dynamics.

\paragraph{Temporal autocovariance.}
For a stationary scalar reset-free model $V_t=\lambda V_{t-1}+gX_t$, use
$g$ for the scalar input gain to distinguish it from confidence-window length.
Assuming finite second moments and convergence of the covariance sum,
\begin{equation}
 \operatorname{Var}(V)=\frac{g^2}{1-\lambda^2}
 \left[\Gamma_X(0)+2\sum_{j\geq1}\lambda^j\Gamma_X(j)\right].
 \label{eq:temporal-covariance}
\end{equation}
Consequently,
\begin{equation}
 \Delta\operatorname{Var}(V)=\frac{g^2}{1-\lambda^2}
 \left[\Delta\Gamma_X(0)+2\sum_{j\geq1}\lambda^j\Delta\Gamma_X(j)\right].
 \label{eq:temporal-covariance-change}
\end{equation}
The zero-lag term vanishes only when marginal input variance is preserved.
The benchmark's temporal jitter can truncate activity at bin boundaries,
so this simplification is not assumed. The sign of the variance change depends
on both marginal variance and temporal correlation changes.

\paragraph{Quadratic deviations: expansive and contracting shifts.}
For a random vector $Y_\gamma$ with mean $\mu_0+\gamma b$ and covariance
$\Sigma_0+\gamma A$, and a fixed positive-definite precision $Q$,
\begin{equation}
 \mathbb E[(Y_\gamma-\mu_0)^\top Q(Y_\gamma-\mu_0)]
 =\operatorname{tr}(Q\Sigma_0)+\gamma\operatorname{tr}(QA)
   +\gamma^2b^\top Qb.
 \label{eq:quadratic-response}
\end{equation}
This expectation is non-decreasing for $\gamma\geq0$ when $A\succeq0$.
That condition describes expansive covariance shifts and is not assumed
for spatial dropout or under-rate corruption. With covariance contraction,
the quadratic distance can decrease. B2 and the dispersion part of B3 use
absolute deviations from clean-calibration distance medians to admit either
tail as evidence. This design does not itself guarantee detection or
monotonicity, and the quadratic expectation result is not a theorem about
the nonlinear descriptors or fused MDD score.

\subsection{Membranes, Spikes, and Spatial Pooling}
\label{app:data-processing}

Let $C$ denote a corruption variable, $U$ the complete pre-threshold membrane
trajectory, and $S$ its spike train. With fixed thresholds and neuron rules,
$S$ is a deterministic function of $U$, so the data-processing inequality gives
\begin{equation}
 I(C;S)\leq I(C;U).
\end{equation}
This statement concerns the complete pre-threshold trajectory. It does not
imply that the compressed descriptors $\varphi$ or $\varphi_{\rm spatial}$
retain more corruption information than spike descriptors, nor that stored
post-reset states alone determine the spikes. Those comparisons require
empirical evaluation. Membrane moments provide a complementary description
of continuous state changes that spike counts can discard.

\paragraph{Spatial information under global average pooling.}
If two moment maps $M,M'$ satisfy
\begin{equation}
 \operatorname{GAP}(M)=\operatorname{GAP}(M'),
\end{equation}
any statistic depending only on that pooled value assigns them the same
value. In particular, spatial permutations preserve GAP while potentially
changing local flatness and persistence. If this equality holds for every
channel and retained moment map, the full pooled descriptor is identical.
This establishes a limitation of pooling, rather than undetectability of
a particular corruption: MDD also retains spatial dispersion and organization.

\subsection{Scope and Limitations}
\label{app:theory-limitations}

The analytical results assume stable linear dynamics, matched initial states,
and the specified noise or stationarity conditions. Reset disagreement limits
the accuracy of the reset-free approximation, especially under severe shifts.
The expansive-covariance result does not cover subtractive corruptions, and
the information inequality applies to complete trajectories rather than their
compressed descriptors. These results motivate the monitored statistics;
they do not prove monotonicity, corruption independence, or performance of
the three-branch detector. Detection performance is measured empirically on
the held-out validation split.

\section{Gen1-C corruption design, reproducibility, and diversity}
\label{app:corruptions}

\paragraph{Purpose and construction.}
Gen1-C is intended to probe distinct acquisition axes rather than to reproduce a single camera model.  We apply one corruption family to an entire sequence, preserving the sequence identity and clean/corrupted pairing.  The six families are: fixed hot pixels (persistent additive offsets), event floods (localized space--time bursts), temporal jitter (independent shifts of the ON and OFF time-bin blocks), event-rate shift (global under- or over-scaling), polarity flip (ON/OFF exchange on a subset of frames), and spatial dropout (persistent rectangular zero regions).  We adopt the corruption-benchmark convention of evaluating predefined families across increasing severity levels~\citep{hendrycks2019imagenetc}. Event-camera literature documents contrast-threshold mismatch~\citep{wang2019threshold}, bandwidth and noise effects~\citep{hu2021v2e}, and background activity~\citep{guo2023badenoising}; \citet{gallego2022survey} review event-camera operating characteristics.

\begin{table}[H]
\centering\small
\caption{Gen1-C parameter schedules.}
\label{tab:gen1c-schedule}
\resizebox{0.95\linewidth}{!}{%
\begin{tabular}{lccccc}
\toprule
Family & S1 & S2 & S3 & S4 & S5\\
\midrule
Hot pixels $(n,\delta)$ & $(10,20)$ & $(30,40)$ & $(80,80)$ & $(150,140)$ & $(300,200)$\\
Event flood $(b,f,\delta)$ & $(2,.05,60)$ & $(4,.08,100)$ & $(8,.12,150)$ & $(15,.18,200)$ & $(25,.25,255)$\\
Temporal jitter $s_{\max}$ & 1 & 2 & 3 & 5 & 8\\
Rate shift $(u,o)$ & $(.80,1.20)$ & $(.65,1.40)$ & $(.50,1.65)$ & $(.35,2.00)$ & $(.20,2.50)$\\
Polarity flip $p$ & .05 & .10 & .20 & .35 & .50\\
Spatial dropout $(k,f_h,f_w)$ & $(1,.05,.05)$ & $(1,.10,.10)$ & $(2,.10,.10)$ & $(2,.15,.15)$ & $(3,.20,.20)$\\
\bottomrule
\end{tabular}}
\end{table}

Here $b$ is the number of burst centres, $f$ the affected image-area fraction, and each event-flood burst spans its centre frame and two neighbours on either side.  The rate-shift generator samples under- versus over-rate direction with the sequence-level generator; both directions are retained in the analysis.  All transformations operate on the $(20,H,W)$ event-count tensor (ten ON and ten OFF bins), clip to the valid count range, and leave the clean tensor untouched.

\begin{figure}[H]
  \centering
  \includegraphics[width=0.99\linewidth]{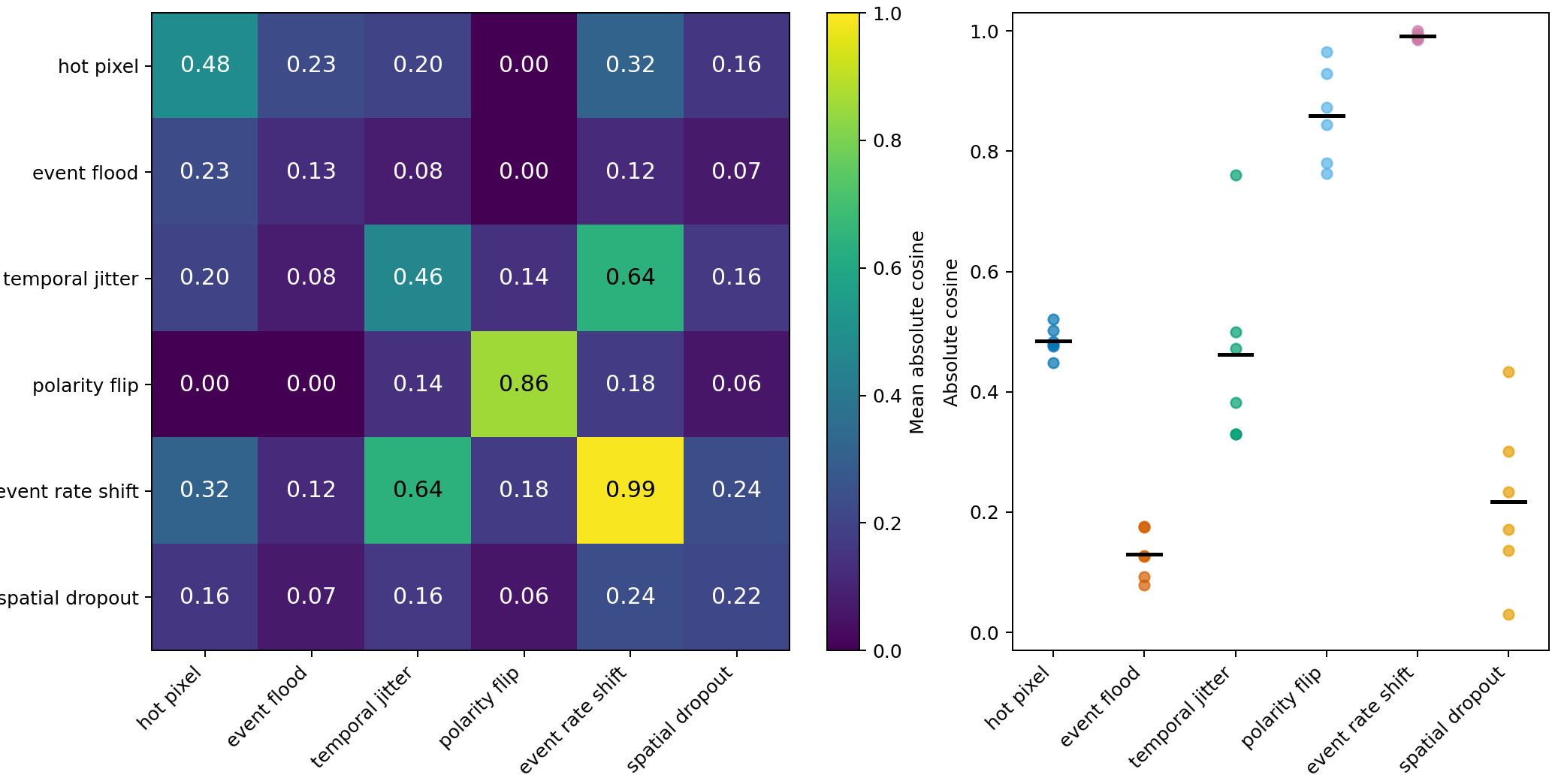}
  \caption{S5 input residual alignment across corruptions.}
  \label{fig:corruption-input-full}
\end{figure}

\paragraph{Do the corruptions produce distinct input changes?}
We compare the transformations before feature extraction, then examine their effects in the descriptor separately.
On six spread-across-pool, standard-length final-pool recordings, we apply each S5 corruption with three independent seeds. We subtract the clean event-count histogram, sum the signed residual into a common time--space--polarity grid,
and compute absolute cosines for matched recordings and seeds. The diagonal in Fig.~\ref{fig:corruption-input-full} compares different seeds of the same family, providing a reference for the off-diagonal cross-family values.

On these six recordings, cross-family input overlap is generally low: the median pairwise mean absolute cosine is $0.16$, and 14 of 15 family pairs are at or below $0.32$. Temporal jitter and rate shift are the clear exception ($0.64$). This pattern is consistent with the operators: hot pixels add counts at fixed locations across time, whereas event flood adds transient patches; temporal jitter moves counts between time bins, polarity flip exchanges ON and OFF channels, rate shift scales counts globally, and spatial dropout removes counts in fixed regions. The suite therefore covers mostly non-overlapping ways of perturbing an event histogram.

\paragraph{What does the descriptor retain?}
For recording $i$ and corruption $c$, let $\delta_{c,i}=(\bar\varphi_{c,i}-\bar\varphi_{0,i})/\sigma_{0,\mathrm{fit}}$, where $\sigma_{0,\mathrm{fit}}$ is the featurewise standard deviation of
clean fitting-pool recording means. We report two different properties of these shifts: their pairwise direction (the mean absolute cosine over matched recordings) and their size (the median $\|\delta_{c,i}\|_2$).

Undefined zero-residual directions are excluded from the descriptor matrix; its diagonal is set to one for display.

\begin{figure}[H]
  \centering
  \includegraphics[width=.62\linewidth]{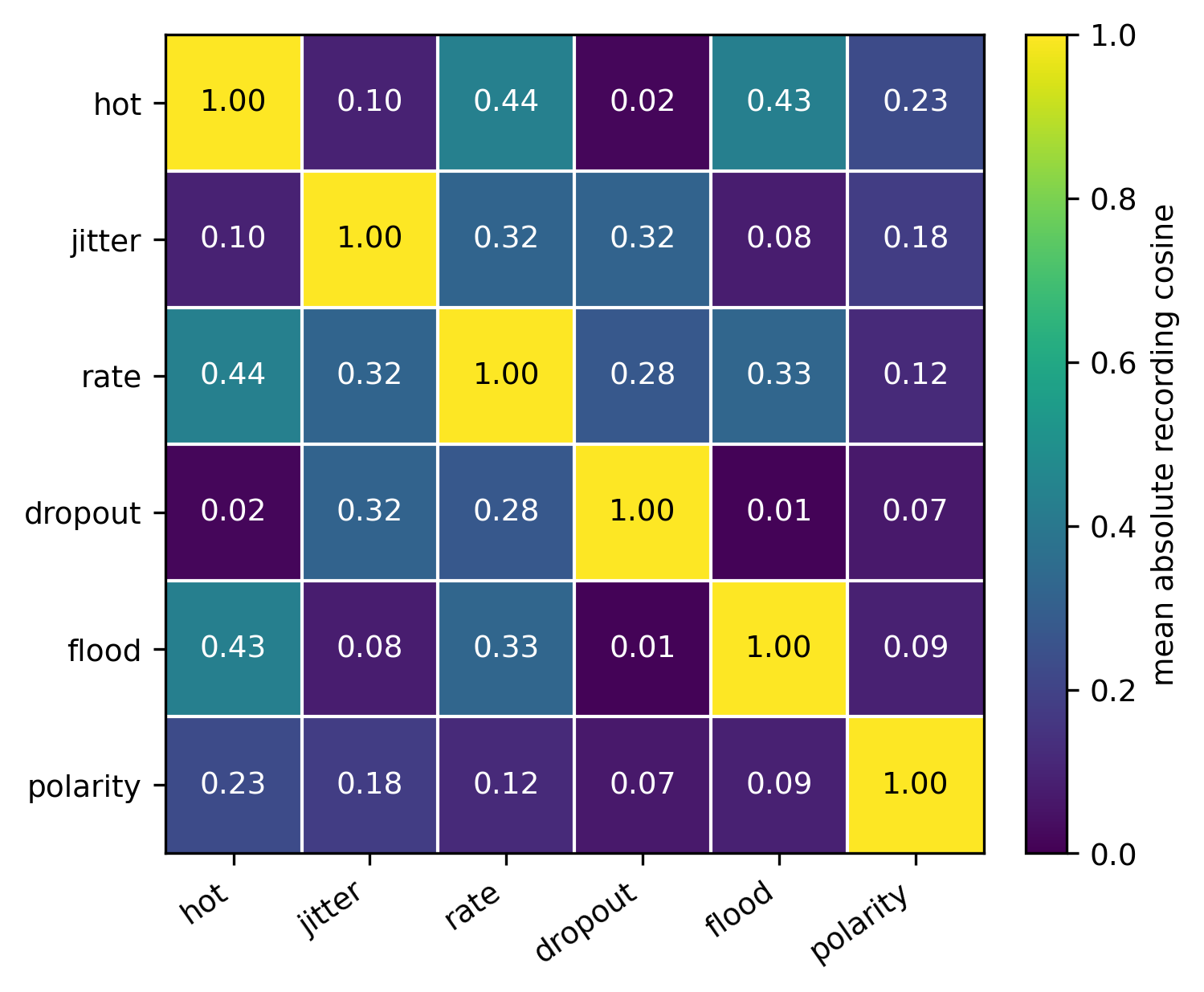}
  \caption{S5 spatial descriptor residual direction alignment.}
  \label{fig:corruption-descriptor-direction}
\end{figure}

The descriptor dimensions differ, so the norm columns describe within-representation scale.

\begin{table}[H]
\centering\small
\caption{Median S5 descriptor residual norms.}
\label{tab:corruption-descriptor-magnitude}
\begin{tabular}{lrr}
\toprule
Corruption & $\varphi$ & $\varphi_{\rm spatial}$\\
\midrule
Hot pixels & 542.4 & $1.53\times10^{5}$\\
Event flood & 578.2 & $4.01\times10^{5}$\\
Temporal jitter & 39.3 & 36.0\\
Polarity flip & 8.1 & 6.3\\
Rate shift & 100.4 & 89.3\\
Spatial dropout & 17.2 & 25.6\\
\bottomrule
\end{tabular}
\end{table}

Let us compare the input and descriptor signals. Interestingly, Feature extraction reverses some overlap patterns. Temporal jitter and rate shift, the strongest input match, become less aligned ($0.67\to0.33$), while the weaker hot-pixel/flood and jitter/dropout matches become stronger ($0.24\to0.43$ and $0.19\to0.44$). Thus, distinct input mechanisms can leave similar signatures for a detector using $\varphi_{\rm spatial}$, whereas strongly aligned input changes need not remain strongly aligned after encoding.

\section{MDD sensitivity and implementation ablations}
\label{app:mdd-sensitivity}

\subsection{Residual projection sensitivity}
For B1, we scale the pooled descriptor using clean-fit statistics, subtract its clean-fit mean, and take the Euclidean norm outside the first $k$ clean principal components. Larger $k$ removes more clean subspace directions. The candidate sweep kept B2, B3, clean fitting, and calibration fixed. Its objective was mean $W=64$ AUROC over the six families and five severities (30 cells) on the sequence-disjoint sensitivity split.

\begin{table}[H]
\centering\small
\caption{MDD width sensitivity.}
\label{tab:mdd-residual-sensitivity}
\begin{tabular}{lrrrr}
\toprule
Removed PCA components & 16 & 32 & 64 & 128\\
\midrule
Mean $W=64$ AUROC & .7916 & .7958 & .7964 & .7949\\
\bottomrule
\end{tabular}
\end{table}

The four widths differ by less than .005 mean AUROC. Width 64 ranks first among these settings, but exceeds width 32 by only .0005; the experiment does not identify a sharply optimal dimension. In the same candidate sweep, a direct clean-scaled descriptor norm scores .7510, and omitting B1 scores .7788, with the other branches fixed.

\subsection{Window-length sensitivity}
The selected model's S5 macro AUROC increases with the observation window (Table~\ref{tab:mdd-residual-window}). This is an accuracy--observation-time trade-off: $W=64$ spans approximately 3.2 seconds at 50 ms per frame.

\begin{table}[H]
\centering\small
\caption{MDD S5 AUROC by window.}
\label{tab:mdd-residual-window}
\setlength{\tabcolsep}{3pt}
\begin{tabular}{lrrrrrrrr}
\toprule
Corruption & 1 & 8 & 16 & 32 & 64 & 128 & 256 & Full\\
\midrule
Hot pixels & 1.000 & 1.000 & 1.000 & 1.000 & 1.000 & 1.000 & 1.000 & 1.000\\
Temporal jitter & .863 & .866 & .870 & .875 & .886 & .899 & .925 & .964\\
Rate shift & .931 & .935 & .941 & .947 & .955 & .963 & .973 & .984\\
Spatial dropout & .963 & .965 & .974 & .980 & .987 & .988 & .993 & .989\\
Event flood & .551 & .614 & .677 & .778 & .892 & .981 & .999 & 1.000\\
Polarity flip & .545 & .560 & .563 & .565 & .568 & .571 & .577 & .593\\
\bottomrule
\end{tabular}
\end{table}

\subsection{Fusion-coefficient sensitivity}

The current detector uses plain maximum fusion. To test whether a less OR-like combiner is preferable, we evaluated
\[
 s_{\alpha}=\max_i s_i-\alpha\,\operatorname{median}_{j\ne i^*}s_j,
 \qquad i^*=\arg\max_i s_i,
\]
after branch standardization. The following table uses S5 final-set windows, $W=64$, and is not used to select the headline MDD numbers.

\begin{table}[H]
\centering
\caption{Fusion-coefficient sweep at S5 and $W=64$.}
\label{tab:mdd-alpha-sensitivity}
\scriptsize
\setlength{\tabcolsep}{3.5pt}
\begin{tabular}{lrrrrrrrr}
\toprule
Corruption & $\alpha=0$ & .25 & .50 & .75 & 1.00 & 1.25 & 1.50 & 2.00\\
\midrule
\texttt{hot\_pixel} & 1.000 & 1.000 & 1.000 & 1.000 & 1.000 & 1.000 & 1.000 & 1.000\\
\texttt{temporal\_jitter} & .886 & .882 & .876 & .865 & .846 & .810 & .742 & .501\\
\texttt{event\_rate\_shift} & .955 & .957 & .956 & .953 & .948 & .941 & .926 & .851\\
\texttt{spatial\_dropout} & .987 & .982 & .966 & .953 & .943 & .933 & .922 & .897\\
\texttt{event\_flood} & .892 & .892 & .893 & .893 & .893 & .893 & .893 & .893\\
\texttt{polarity\_flip} & .568 & .564 & .560 & .556 & .551 & .544 & .529 & .496\\
\textbf{Mean} & .881 & .880 & .875 & .870 & .863 & .854 & .835 & .773\\
\bottomrule
\end{tabular}
\end{table}

Subtracting the median evidence helps rate shift slightly near $\alpha=.25$, but progressively suppresses temporal jitter and spatial dropout. The plain maximum remains the strongest macro choice at S5 for MDD. We also tested alternative joins of the calibrated branch scores: arithmetic mean, top-two mean, noisy-OR,
percentile-of-maximum, percentile-of-mean, chi-square combination, and individual-branch baselines. The compact summary below reports S5, $W=64$ values over the six families; maximum fusion is the strongest overall option.

\begin{figure}[H]
    \centering
    \begin{minipage}[c]{.38\linewidth}
      \centering
      \includegraphics[width=\linewidth]{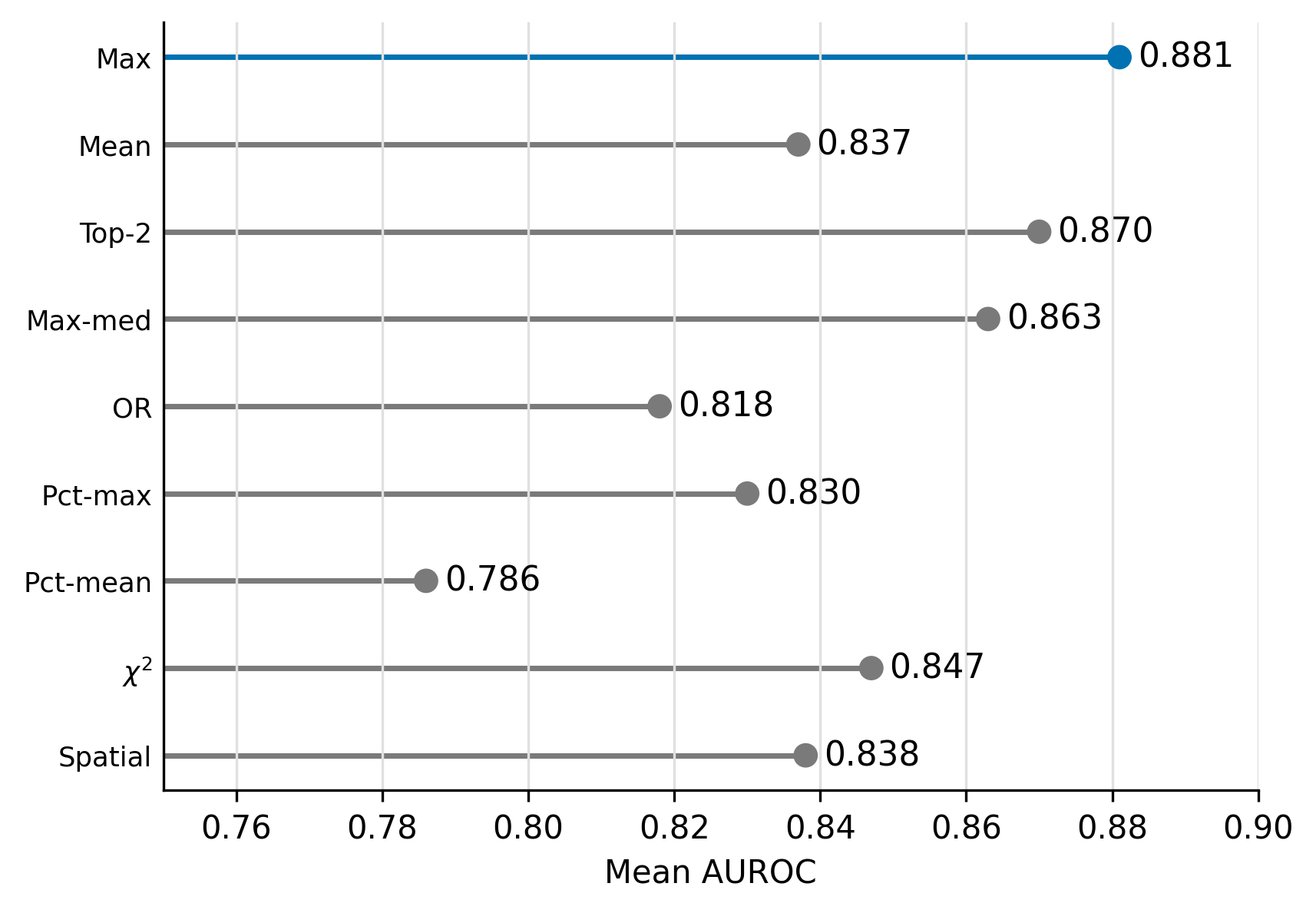}
    \end{minipage}\hfill
    \begin{minipage}[c]{.6\linewidth}
      \centering
      \includegraphics[width=\linewidth]{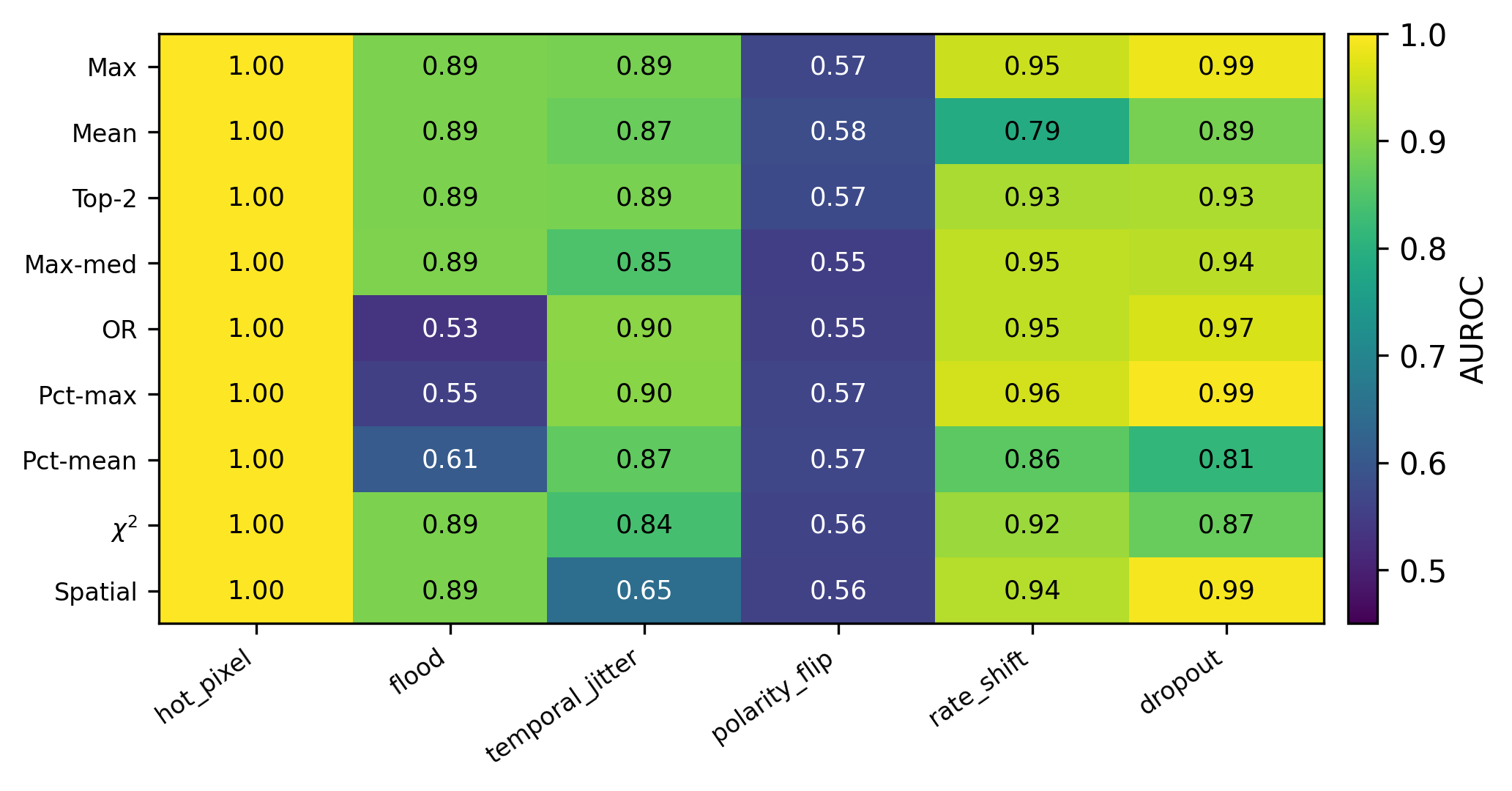}
    \end{minipage}
    \caption{MDD combiner results at S5 and $W=64$.}
    \label{fig:mdd-combiner-sensitivity}
  \end{figure}

In the median-subtracted rule, selecting its $\alpha$ from corrupted final labels would make the selection corruption-informed. We therefore treat the $\alpha$ sweep and the alternative joins as diagnostic ablations. A deployed corruption-blind choice must be fixed a priori or selected using a separate development split, then evaluated once on held-out corrupted recordings.

\subsection{Softmax-style smoothing check}

As a separate numerical check, we replaced the maximum by a temperature-controlled soft maximum. At S5 and $W=64$, the mean AUROC was .881, .881, .879, .874, .865, .855, and .846 for temperatures $\tau=0.1,0.25,0.5,1,2,4,8$ (rounded). The small differences near $\tau=0$ do not justify an additional temperature parameter.

\section{Cross-dataset and cross-model evaluation}
\label{app:generalization}

The Gen1 result uses the 240$\times$304 Prophesee sensor, ten 5-ms bins, and the previously described hybrid PLIF-front-end/ANN detection model.  We use Gen4 and SpikeYOLO because they change different sources of variation: Gen4
changes the sensor and dataset, while SpikeYOLO changes the host network. Each transfer run refits and recalibrates MDD on its own clean data; no Gen1 references or thresholds are reused.

\paragraph{Transfer to Prophesee Gen4}

Gen4 uses the Prophesee 1-Mpixel, 1280$\times$720 sensor~\citep{perot2020megapixel} and a similar model to the one we tested on Gen1.  Its input is the same stacked-histogram family but with $T=10$ bins over 50 ms, Gen4's three retained classes (pedestrian, two-wheeler, car), and the Gen4 preprocessing downsample-by-two path.  The
loaded detector uses the V1-attention backbone at $(384,640)$, a depth-.33 PAFPN with stages 2/3/4, and a three-class detection head; its evaluation configuration uses five-frame sequences.  This differs from Gen1 in sensor
resolution, class set, scene distribution, checkpoint, and detector input geometry.

\paragraph{Cross-model evaluation with SpikeYOLO}

The second host is the downloaded SpikeYOLO detector~\citep{luo2024spikeyolo}, with a P3--P5 detection pyramid and a two-step temporal simulation in our evaluation configuration.  Its convolutional blocks use \texttt{MultiStepLIFNode} neurons throughout the spiking backbone and detection pathway.  This differs from the SNN--ANN detector used in the main Gen1 experiments: the latter exposes four PLIF layers
($64/128/256/256$ channels) from a spiking front end and then passes their features to a conventional ANN YOLOX/PAFPN detection head.  SpikeYOLO instead keeps the feature extraction and detection pathway spiking and uses a shorter temporal simulation.

\begin{table}[H]
  \caption{Host-specific MDD AUROC at $W=64$ by severity.}
  \label{tab:mdd-transfer-auroc}
  \begin{minipage}[t]{0.49\linewidth}
  \centering\small
  {\footnotesize\textbf{(a) Gen4}}\\[2pt]
  \resizebox{\linewidth}{1.3 cm}{%
  \begin{tabular}{lccccc}
  \toprule
  Corruption & 1 & 2 & 3 & 4 & 5\\
  \midrule
  \texttt{hot\_pixel} & .994 & 1.000 & 1.000 & 1.000 & 1.000\\
  \texttt{temporal\_jitter} & .568 & .593 & .607 & .631 & .650\\
  \texttt{event\_rate\_shift} & .498 & .510 & .685 & .757 & .840\\
  \texttt{spatial\_dropout} & .501 & .510 & .519 & .650 & .789\\
  \texttt{event\_flood} & .572 & .625 & .662 & .770 & .874\\
  \texttt{polarity\_flip} & .500 & .501 & .501 & .502 & .503\\
  \bottomrule
  \end{tabular}}
  \end{minipage}\hfill\begin{minipage}[t]{0.49\linewidth}
  \centering\small
  {\footnotesize\textbf{(b) SpikeYOLO}}\\[2pt]
  \resizebox{\linewidth}{1.3 cm}{%
  \begin{tabular}{lccccc}
  \toprule
  Corruption & 1 & 2 & 3 & 4 & 5\\
  \midrule
  \texttt{hot\_pixel} & .540 & .450 & .600 & .710 & .830\\
  \texttt{temporal\_jitter} & .750 & .960 & .670 & .790 & .940\\
  \texttt{event\_rate\_shift} & .630 & .780 & .830 & .830 & .830\\
  \texttt{spatial\_dropout} & .570 & .560 & .570 & .450 & .770\\
  \texttt{event\_flood} & .540 & .570 & .560 & .560 & .670\\
  \texttt{polarity\_flip} & .775 & .815 & .830 & .840 & .930\\
  \bottomrule
  \end{tabular}}
  \end{minipage}
  \end{table}

The results show that MDD transfers across both a sensor/dataset shift and a host-model change without reusing Gen1 references. However, the corruption ranking changes substantially between Gen4 and SpikeYOLO. This is expected: MDD scores deviations in the host’s internal activity, so sensitivity depends on the full deployed pipeline; sensor resolution and event statistics, histogram construction and resizing, temporal binning, preprocessing, network architecture, neuron dynamics, and the monitored layers; not solely on the input-space corruption magnitude. For example, polarity flip is near chance on Gen4 but rises to .930 on SpikeYOLO at S5, whereas spatial dropout is stronger on Gen4 compared to SpikeYOLO. Thus, MDD should be fitted, calibrated, and evaluated for each concrete sensor--preprocessing--model stack; the transferable element is the clean-only fitting procedure, not a universal ordering of corruption difficulty.

\section{Efficient MDD: descriptor, scorer, AUROC, and MCU results}
\label{app:efficient-mdd-app}

Efficient MDD  monitors four frozen channels in each of the four PLIF layers and samples every eighth spatial
position in layer 1 and every fourth position in layers 2--4. The sampled post-step Vmem tensors have shapes $(10,4,16,20)$, $(10,4,16,20)$, $(10,4,8,10)$, and $(10,4,8,10)$. The channel indices and sampling grid are
fixed using clean-fit data. Input collection from the host SNN and transfer to the monitor are outside the measured C compute interval.

The extractor accumulates temporal mean, variance, and kurtosis maps, their spatial summaries, and four layerwise persistent-flatness fractions. The scorer uses 84 active features: 80 moment/spatial features and four
persistence features. It takes the maximum of five clean-calibrated scores: four layerwise quadratic-form deviations and one persistence deviation. It uses no PCA projection, RCF lookup, or stored clean reference set.
Persistent counters are reset at recording boundaries; averaging over a 64-frame decision window is performed outside the per-frame scorer.

\begin{figure*}[t]
\centering
\resizebox{\linewidth}{!}{%
\begin{tikzpicture}[font=\small,>=Latex,
 box/.style={draw,rounded corners,align=center,fill=blue!5,inner sep=5pt},
 branch/.style={box,fill=green!7,text width=2.1cm},
 ext/.style={box,fill=gray!10},
 flow/.style={->,thick}]
\node[ext] (source) at (5.4,0.5)
 {SNN post-step Vmem};
\node[box] (input) at (5.4,-1.3)
 {Sampled Vmem};
\node[box] (maps) at (5.4,-2.8)
 {Shared statistics};
\draw[flow] (source) -- (input);
\draw[flow] (input) -- (maps);
\foreach \x/\l in {0/1,2.7/2,5.4/3,8.1/4}{
 \node[branch] (phi\l) at (\x,-4.4)
 {Layer $\l$: $\phi_\l$\\};
 \node[branch] (score\l) at (\x,-6.1)
 {Mahalanobis $\l$\\};
 \draw[flow] (maps.south) -- (5.4,-3.5) -| (phi\l.north);
 \draw[flow] (phi\l) -- (score\l);
}
\node[branch,fill=orange!12] (persist) at (10.8,-4.4)
 {Patch flatness};
\node[branch,fill=orange!12] (score5) at (10.8,-6.1)
 {Spatial Branch};
\draw[flow] (maps.south) -- (5.4,-3.5) -| (persist.north);
\draw[flow] (persist) -- (score5);
\draw[flow] (persist.east) -- (12.4,-4.4) -- (12.4,-3.4)
 -- (10.8,-3.4) -- (persist.north);
\node[anchor=west,align=left,font=\scriptsize] at (12.5,-3.9)
 {Counts + frame index\\retained for next frame};
\node[box] (fusion) at (5.4,-8.0)
 {Maximum of five calibrated scores};
\foreach \l in {1,2,3,4,5}{
 \draw[thick] (score\l.south) -- (score\l.south |- 0,-7.15);
}
\draw[thick] (0,-7.15) -- (10.8,-7.15);
\draw[flow] (5.4,-7.15) -- (fusion.north);
\draw[dashed,gray,rounded corners] (-1.45,-0.5) rectangle (15.4,-8.7);
\node[anchor=north east,font=\footnotesize,align=right] at (15.2,-0.65)
 {Measured native-C region\\Sampled Vmem $\rightarrow$ score};
\end{tikzpicture}}
\caption{Efficient MDD}
\label{fig:efficient-mdd-flow}
\end{figure*}
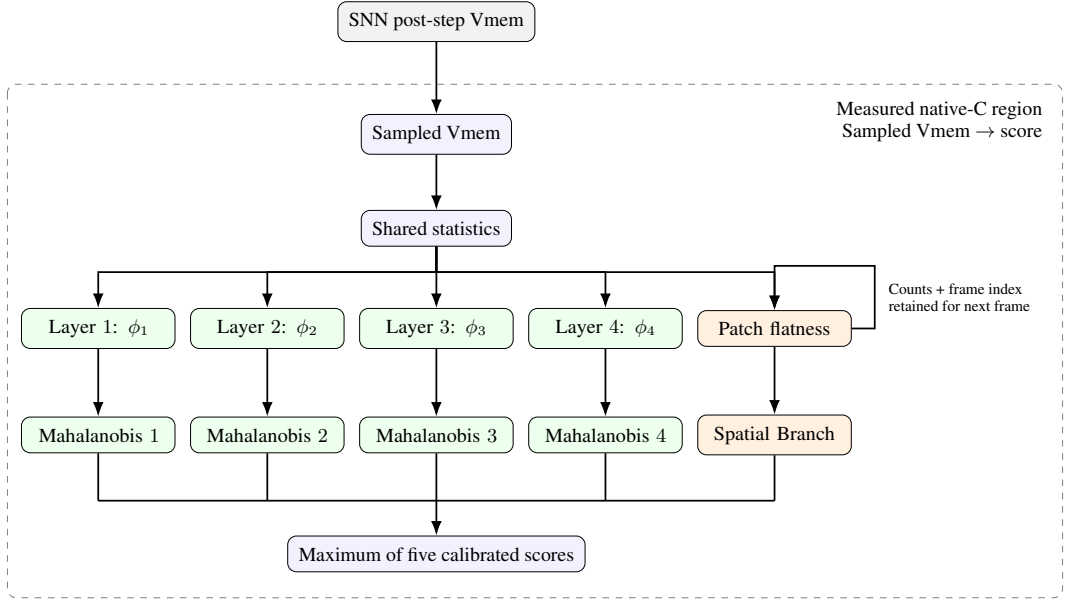

\subsection{AUROC by severity}

\begin{table}[H]
\centering\footnotesize
\caption{MDD and Efficient MDD AUROC.}
\label{tab:efficient-mdd-current-auroc}
\resizebox{\linewidth}{!}{%
\begin{tabular}{l*{10}{c}}
\toprule
& \multicolumn{5}{c}{MDD} & \multicolumn{5}{c}{Efficient MDD}\\
Corruption & S1 & S2 & S3 & S4 & S5 & S1 & S2 & S3 & S4 & S5\\
\midrule
Hot pixel & 1.000 & 1.000 & 1.000 & 1.000 & 1.000 & .571 & .908 & 1.000 & 1.000 & 1.000\\
Temporal jitter & .692 & .745 & .798 & .847 & .886 & .559 & .604 & .631 & .659 & .748\\
Event rate shift & .519 & .579 & .888 & .906 & .955 & .508 & .515 & .728 & .743 & .888\\
Spatial dropout & .800 & .952 & .962 & .984 & .987 & .493 & .493 & .500 & .505 & .926\\
Event flood & .556 & .602 & .692 & .795 & .892 & .545 & .615 & .710 & .791 & .889\\
Polarity flip & .511 & .520 & .534 & .553 & .568 & .503 & .493 & .500 & .495 & .534\\
\bottomrule
\end{tabular}}
\end{table}

The efficiency reduction preserves the large, localized deviations caused by hot pixels and event flood, and retains strong S5 detection for rate shift and spatial dropout. Its loss is concentrated in temporal jitter ($-.138$ AUROC at S5) and, to a lesser extent, rate shift and spatial dropout, suggesting that the reduced channel and spatial sampling discards some of the fine spatiotemporal evidence used by the full descriptor. Compared with MDD, this degradation is more pronounced at lower severities: the reduced descriptor reaches high AUROC later for temporal jitter, rate shift, spatial dropout, and event flood. Hot pixels are the exception, remaining easy to detect from S2, while polarity flip is difficult for both detectors throughout.

\subsection{Physical-MCU measurements}
Table~\ref{tab:efficient-mdd-current-mcu}reports the maximum observed sampled-Vmem-to-score time across 330 checked measurements per board. The compute interval includes four layer extraction/state calls and scoring, but excludes the external SNN, raw-Vmem gathering, and fixture staging. The implementation is a static float32 C routine with fixed channel indices, fixed spatial strides, and clean-fitted scorer parameters compiled into read-only arrays. It processes each sampled Vmem layer in place, accumulates the required temporal moments and persistence state without materializing full descriptors, then evaluates the five fixed score branches and their maximum fusion.

\begin{table}[H]
\centering\small
\caption{Efficient MDD physical-MCU timing.}
\label{tab:efficient-mdd-current-mcu}
\setlength{\tabcolsep}{8pt} 
\begin{tabular}{llrr}
\toprule
Board / MCU & Core / FP & MHz & Captured max (ms)\\
\midrule
B-L475E-IOT01A / STM32L475VG & M4F / hardware & 80 & 10.8388\\
nRF52840 DK / nRF52840 & M4F / hardware & 64 & 13.6235\\
nRF52832 DK / nRF52832 & M4F / hardware & 64 & 19.4250\\
DWM3001 / nRF52833 & M4F / hardware & 64 & 19.4555\\
STM32F103RE & M3 / software & 72 & 247.0523\\
\bottomrule
\end{tabular}
\end{table}

\begin{table}[H]
\centering
\small
\caption{Hardware resource requirements for MDD implementation.}
\label{tab:mdd-resources}
\begin{tabular}{lrl}
\toprule
Metric & Value & Notes / Scope \\
\midrule
Persistent state memory & 5.36 KB & Monitor state \\
Sampled-layer buffer & 51.20 KB & Peak single-layer input \\
Total monitor RAM & 56.56 KB & Excludes call stack \\
\midrule
Input traffic & 32,000 floats/frame & 128 KB read per frame \\
Scorer coefficient products & 844 ops/frame & Triangular forms and flatness branch \\
Temporal-moment cells & $\le$ 3,200/frame & Upper bound (data-dependent) \\
Flatness patches & $\le$ 4,304/frame & Upper bound (data-dependent) \\
\bottomrule
\end{tabular}
\end{table}

\section{Why Polarity Flip Is Difficult to Detect}
\label{app:polarity-failure}

Polarity flip is the weakest corruption for MDD on the primary Gen1 host:
at S5 its AUROC is $0.545$ per frame and $0.568$ at $W=64$.
The diagnostics below identify two sources of difficulty: the corruption
leaves some frames unchanged, and moment aggregation can discard
polarity-sensitive membrane changes. They also distinguish this detection
failure from the host detector's relatively small aggregate task degradation.

\subsection{Corruption Rule and Unchanged Frames}

The binned corruption exchanges the complete ON and OFF ten-bin blocks
independently for each frame, with probabilities
$q=0.05,0.10,0.20,0.35,0.50$ at S1--S5. It does not independently reverse
each event's polarity. A selected frame retains its total activity and
spatial support while exchanging its polarity channels; an unselected frame
is identical to its clean counterpart.

For a fixed memoryless frame score, let $A_{\rm full}$ be its population
AUROC against inputs that are always swapped. Independence of the flip mask
gives the tie-aware mixture identity
\begin{equation}
 A_q=(1-q)\frac12+qA_{\rm full}
     =\frac12+q\left(A_{\rm full}-\frac12\right).
 \label{eq:polarity-mixture}
\end{equation}
Even perfect separation of fully swapped frames therefore gives
$A_q\leq\frac12+\frac q2$, which is $0.75$ at S5. This population identity
isolates dilution by untouched frames. It is not a ceiling for the deployed
window score or the history-dependent spatial branch: both can accumulate
evidence across frames. The observed weak detection also requires examining
what evidence survives feature extraction.

\subsection{Membrane Changes Can Cancel in the Descriptor}

Polarity-sensitive activity need not disappear from the network to disappear
from its moment descriptor. At a fixed layer, channel, and spatial location,
let $v_t$ and $v'_t$ be the clean and fully swapped membrane responses.
Define their even and odd components by
\begin{equation}
 A_t=\frac{v_t+v'_t}{2},\qquad B_t=\frac{v_t-v'_t}{2},\qquad
 a_t=A_t-\mathbb E_t A_t,\quad b_t=B_t-\mathbb E_t B_t,
\end{equation}
where $\mathbb E_t$ is the finite average over temporal bins.
The exact moment differences, with primes denoting swapped responses, are
\begin{align}
 \mu'-\mu&=-2\mathbb E_t B_t,\\
 \nu'-\nu&=-4\mathbb E_t[a_tb_t],\\
 m'_4-m_4&=-8\mathbb E_t[a_t^3b_t+a_tb_t^3].
 \label{eq:polarity-cancellation}
\end{align}
Here $\nu$ is the unfloored central variance and $m_4$ the fourth central
moment. These identities are algebraic and do not assume linear neurons,
small perturbations, or negligible resets. A large odd response $B_t$ can
coexist with small moment changes when its signed contributions cancel.
If all three differences vanish, the implemented variance floor and excess
kurtosis are also unchanged. For approximate equality, the kurtosis
denominator must additionally be controlled.

Global average pooling can cancel further differences across locations.
To quantify this, for a paired moment-difference map $d(c,z)$ we measure
\begin{equation}
 \rho(d)=\frac{\sum_c|\operatorname{GAP}(d_c)|}
 {\sum_c\operatorname{GAP}(|d_c|)}.
 \label{eq:polarity-retention}
\end{equation}
A value near zero indicates strong cancellation of signed changes under
spatial pooling; a value of one indicates no cancellation. This statistic
measures aggregation loss, rather than clean-versus-corrupted separability.

\subsection{Checkpoint Diagnostics}

A pilot evaluated fully swapped real frames from four segments
of one underlying recording, with fresh neuron and spatial state for each
observation. It executed the four monitored PLIF blocks with their checkpoint weights strictly loaded. This isolates descriptor behavior.

\begin{table}[H]
\centering\small
\caption{Polarity changes before and after pooling.}
\label{tab:polarity-pooling}
\begin{tabular}{lrrrr}
\toprule
Diagnostic median & PLIF 1 & PLIF 2 & PLIF 3 & PLIF 4\\
\midrule
Raw membrane relative change & .211 & .257 & .345 & .634\\
Mean-map retention $\rho$ & .237 & .171 & .232 & .195\\
Variance-map retention $\rho$ & .273 & .237 & .260 & .184\\
Kurtosis-map retention $\rho$ & .149 & .112 & .117 & .068\\
\bottomrule
\end{tabular}
\end{table}

The raw membrane relative change is
$\|V'-V\|_2/\|V\|_2$; undefined zero-denominator retention ratios are
excluded. Variance retention uses the unfloored central variance, while
kurtosis uses the implemented variance floor. Substantial membrane changes
coexist with much smaller signed changes after pooling. In the deepest
layer, the median retained fractions are only $19.5\%$ for mean changes,
$18.4\%$ for variance changes, and $6.8\%$ for kurtosis changes.
The pooled descriptor changes on 31 of the 32 frames, so these results
support attenuation of polarity evidence rather than exact invariance on
ordinary inputs. The checkpoint's first-layer ON/OFF filters have relative
difference $1.590$; nearly identical polarity filters do not explain the result.

\paragraph{An exact descriptor-collision mechanism.}
The audited stem applies a shared binwise computation under its extraction
and reset protocol. Construct ten input bins by concatenating five arbitrary
two-polarity bins with their polarity-swapped copies. Swapping ON/OFF then
permutes the two five-bin halves. Any shared nonlinear binwise response
undergoes the same permutation, which preserves its temporal moments at
every spatial location. Consequently, both the pooled descriptor and the
spatial moment maps are invariant. With matched histories and initial
persistence state, the organization features are invariant as well.

In the executed checkpoint witness, raw membrane relative changes across
the four layers are $.717,.648,.485,.441$, whereas the relative changes of
the pooled and freshly initialized spatial descriptors are only
$5.86\times10^{-8}$ and $1.09\times10^{-7}$. This demonstrates a nontrivial
input change that the descriptors cannot distinguish. The changed real
pilot frames do not satisfy this exact permutation relation, so the witness
establishes a possible failure mechanism rather than its frequency in Gen1.
Any fixed MDD scorer assigns equal scores to exactly equal complete
descriptor inputs; changing the downstream score alone cannot resolve such
collisions.

\subsection{What the Current MDD Results Show}

\begin{table}[H]
\centering\small
\caption{Current MDD polarity detection at S5.}
\label{tab:polarity-current-mdd}
\begin{tabular}{lc}
\toprule
Score or aggregation & AUROC\\
\midrule
B1: PCA residual, $W=64$ & .583\\
B2: deep-layer deviation, $W=64$ & .512\\
B3: spatial deviation, $W=64$ & .558\\
Fused MDD, $W=1$ & .545\\
Fused MDD, $W=64$ & .568\\
Fused MDD, full recording & .593\\
\bottomrule
\end{tabular}
\end{table}

No branch provides strong polarity separation. In particular, the spatial
branch's weak result shows that adding organization statistics does not
recover enough evidence for this corruption. Increasing aggregation from
one frame to an entire recording improves AUROC only modestly. Removing B3
raises S5 $W=64$ polarity AUROC to $.598$, but separation remains weak;
maximum fusion contributes to the loss without being its sole cause.
Together with the checkpoint diagnostics, these results support a limitation
of the retained statistics and their clean-fit scoring, rather than a problem
that window averaging alone resolves.

\paragraph{Relation to task performance.}
Polarity flip also causes relatively little aggregate task degradation:
mAP decreases from $.358$ on clean inputs to $.344$ at S5. Nevertheless,
MDD is intended to detect corruption, and its weak polarity separation
remains a limitation. The membrane diagnostics offer a more direct
explanation: polarity-sensitive responses change, but much of their signed
variation cancels in the retained moment statistics. Thus a corruption
can alter internal activity without producing a strong deviation in
the descriptors used by MDD.

\paragraph{Conclusion.}
Polarity exchange preserves activity magnitude and spatial support, while
its signed membrane changes can cancel under temporal moments and spatial
pooling. This leaves MDD with weak evidence even when individual membrane
responses change substantially. Untouched-frame dilution adds difficulty,
and the current branch and window results show that neither spatial
organization nor longer averaging fully addresses it. The findings motivate
testing descriptors that retain polarity-specific or ordered temporal
information.

\end{document}